\documentclass[manuscript,screen,acmlarge]{acmart}

\usepackage{tabularx}
\usepackage{enumitem}
\usepackage{tikz}
\usepackage{xcolor}

\usepackage{subcaption}
\usepackage{mathtools}

\AtBeginDocument{%
  \providecommand\BibTeX{{%
    \normalfont B\kern-0.5em{\scshape i\kern-0.25em b}\kern-0.8em\TeX}}}

\setcopyright{acmcopyright}
\copyrightyear{2022}
\acmYear{2022}
\acmDOI{10.1145/1122445.1122456}
\acmJournal{IMWUT}
\acmVolume{1}
\acmNumber{1}
\acmArticle{1}
\acmMonth{1}
\acmPrice{15.00}
\acmISBN{978-1-4503-1111-1/18/06}

\newcommand{\parjump}{\vspace{+0.5em}}

\begin{document}

\newcommand{\problem}{TSMDS}
\newcommand{\solution}{ColloSSL}

\newcommand{\revisions}[1]{\textcolor{black}{{#1}}}
\newcommand{\red}[1]{\textcolor{red}{{#1}}}


\newcommand*\circled[1]{\tikz[baseline=(char.base)]{
            \node[shape=circle,fill,inner sep=1.5pt] (char) {\textcolor{white}{#1}};}}

\newcommand*\grey[1]{\tikz[baseline=(char.base)]{
            \node[shape=circle,fill={rgb:black,1;white,2},inner sep=1.5pt] (char) {\textcolor{white}{#1}};}}

\newcolumntype{P}[1]{>{\centering\arraybackslash}p{#1}}
\newcommand{\STAB}[1]{\begin{tabular}{@{}c@{}}#1\end{tabular}}
\newcommand{\specialcell}[2][c]{%
  \begin{tabular}[#1]{@{}c@{}}#2\end{tabular}}
  
\title{\solution{}: Collaborative Self-Supervised Learning for Human Activity Recognition}

\author{Yash Jain}
\authornote{Ordered alphabetically, equal contribution. The work was done when the authors were visiting researchers at Nokia Bell Labs, United Kingdom.}
\email{yashjain@gatech.edu}
\affiliation{%
	\institution{Georgia Tech}
	\country{United States of America}
}

\author{Chi Ian Tang}
\authornotemark[1]
\email{cit27@cam.ac.uk}
\affiliation{%
	\institution{University of Cambridge}
	\country{United Kingdom}
}

\author{Chulhong Min}
\email{chulhong.min@nokia-bell-labs.com}
\affiliation{%
	\institution{Nokia Bell Labs}
	\country{United Kingdom}
}
\author{Fahim Kawsar}
\email{fahim.kawsar@nokia-bell-labs.com}
\affiliation{%
	\institution{Nokia Bell Labs}
	\country{United Kingdom}
}

\author{Akhil Mathur}
\email{akhil.mathur@nokia-bell-labs.com}
\affiliation{%
	\institution{Nokia Bell Labs}
	\country{United Kingdom}
}

\renewcommand{\shortauthors}{Jain and Tang, et al.}

\begin{abstract}
A major bottleneck in training robust Human-Activity Recognition models (HAR) is the need for large-scale labeled sensor datasets. Because labeling large amounts of sensor data is an expensive task, unsupervised and semi-supervised learning techniques have emerged that can learn good features from the data without requiring any labels. In this paper, we extend this line of research and present a novel technique called Collaborative Self-Supervised Learning (ColloSSL) which leverages unlabeled data collected from \emph{multiple} devices worn by a user to learn high-quality features of the data. A key insight that underpins the design of \solution{} is that unlabeled sensor datasets simultaneously captured by multiple devices can be viewed as natural transformations of each other, and leveraged to generate a supervisory signal for representation learning. We present three technical innovations to extend conventional self-supervised learning algorithms to a multi-device setting: a \emph{Device Selection} approach which selects positive and negative devices to enable contrastive learning, a \emph{Contrastive Sampling} algorithm which samples positive and negative examples in a multi-device setting, and a loss function called \emph{Multi-view Contrastive Loss} which extends standard contrastive loss to a multi-device setting. Our experimental results on three multi-device datasets show that \solution{} outperforms both fully-supervised and semi-supervised learning techniques in majority of the experiment settings, resulting in an absolute increase of upto 7.9\% in $F_1$ score compared to the best performing baselines. We also show that \solution{} outperforms the fully-supervised methods in a low-data regime, by just using one-tenth of the available labeled data in the best case.
\end{abstract}


\maketitle

\section{Introduction}
\label{sec:intro}
The adoption of human activity recognition (HAR) applications in mobile and wearable devices has increased significantly over the last few decades, owing to the advancements in computational models that process raw sensor data to infer human activities. Typically these computational models are trained using supervised learning, i.e., it requires a set of labeled data samples to train the models. More recently, with the popularity of data-hungry deep learning models in HAR~\cite{hammerla2016deep,lee2017human,ronao2016human}, the need for large-scale labeled training data has become more pronounced. However, prior works have highlighted that collecting large-scale labeled HAR data is cumbersome especially outside of laboratory settings~\cite{chen2021sensecollect}. One key bottleneck for labeling HAR datasets is that sensor streams (e.g., accelerometer traces) are not easy to interpret by visual inspection, which makes any post-hoc labeling efforts non-trivial. A direct consequence of this challenge is that HAR data collection efforts~\cite{anguita2013public, roggen2010collecting, reiss2012introducing, micucci2017unimib} are usually done at a small-scale, in a controlled or semi-controlled setting, and mostly involve less than 50 participants, thereby resulting in models which do not generalize in real-life situations.  

In contrast to the challenges involved in data labeling, the collection of \emph{unlabeled} HAR data is much easier due to the ubiquity of senor-enabled devices (e.g., smartphone, wearables) in our daily lives. As a result, machine learning approaches which utilize unlabeled data during training have been gaining prominence in the HAR literature. One of the most exciting paradigm in this direction has been self-supervised learning (SSL)~\cite{jing2020self}, where the core idea is to leverage the inherent structure present in the (unlabeled) input data to derive a supervisory signal. Typically it is done by defining a \emph{Pretext Task}, wherein a set of pre-defined transformations are applied to the input data and a deep neural network is trained to predict those transformations in the data. For example, in image datasets, the Pretext task could involve rotating an input image by 60\textdegree, and the deep neural network is trained to learn that the original and rotated image share the same embedding. We usually are not interested in the accuracy of the model on the Pretext task. Rather, we care about the intermediate representations (or features) learned by the model, with the expectation that these features will carry good semantic or structural information about the signal, which can be useful for various downstream tasks. 

In recent years, we have seen some exciting explorations of self-supervised learning for activity recognition with inertial sensors. Saeed et al.~\cite{saeed2019multi} explored SSL in a multi-task setting wherein they proposed eight different transformations for accelerometer data, and trained a feature extractor to predict whether each transformation has been applied or not. In doing so, they found that the feature extractor managed to learn the inherent structure in the data and generate good features. Tang et al.~\cite{tang2021selfhar} extended this idea by applying teacher-student training and showed performance gains over prior work. Haresamudram et al.~\cite{haresamudram2020masked} investigated the idea of masking sensor data at certain timesteps and training the feature encoder to reconstruct the missing data. More recently, Haresamudram et al.~\cite{haresamudram2021contrastive} also proposed Contrastive Predictive Coding which leverages the long-term temporal structure of sensor data streams for self-supervised learning. SSL has also been used in the context of change point detection~\cite{deldari2021time} and federated learning~\cite{saeed2020federated}.

\begin{figure}[t]
\subfloat[On-body inertial sensors]{\label{tsmds-a}\includegraphics[width=0.33\linewidth]{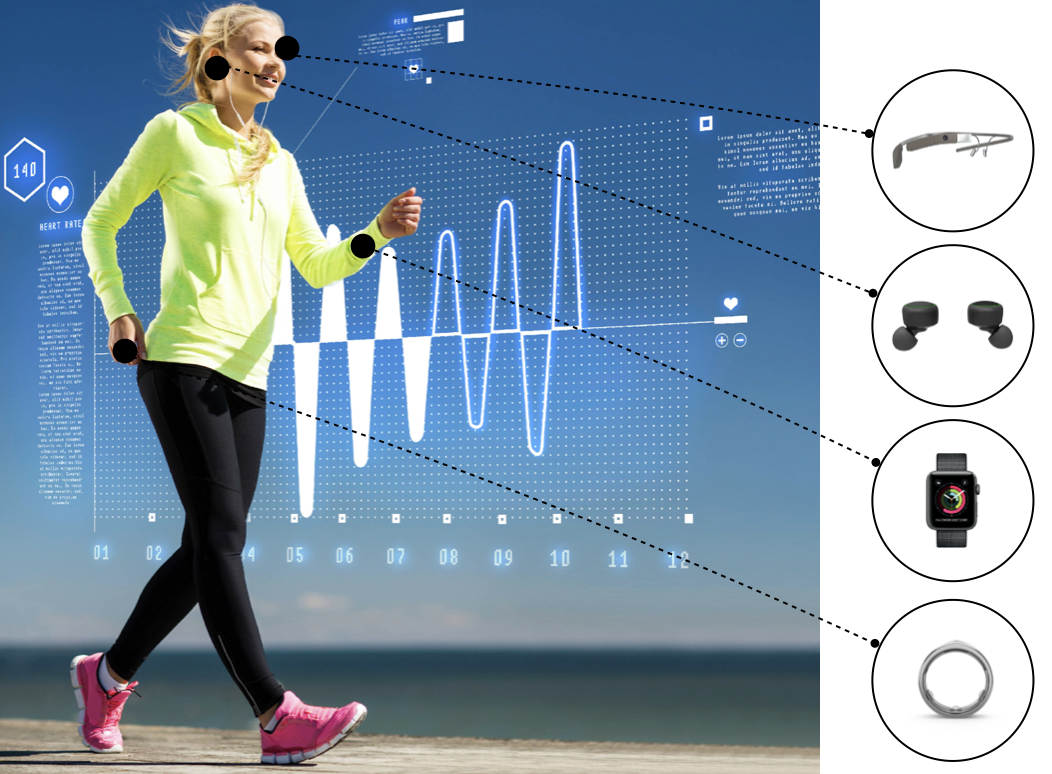}}\hfill
\subfloat[Acoustic sensors in a living space]{\label{tsmds-b}\includegraphics[width=0.33\linewidth]{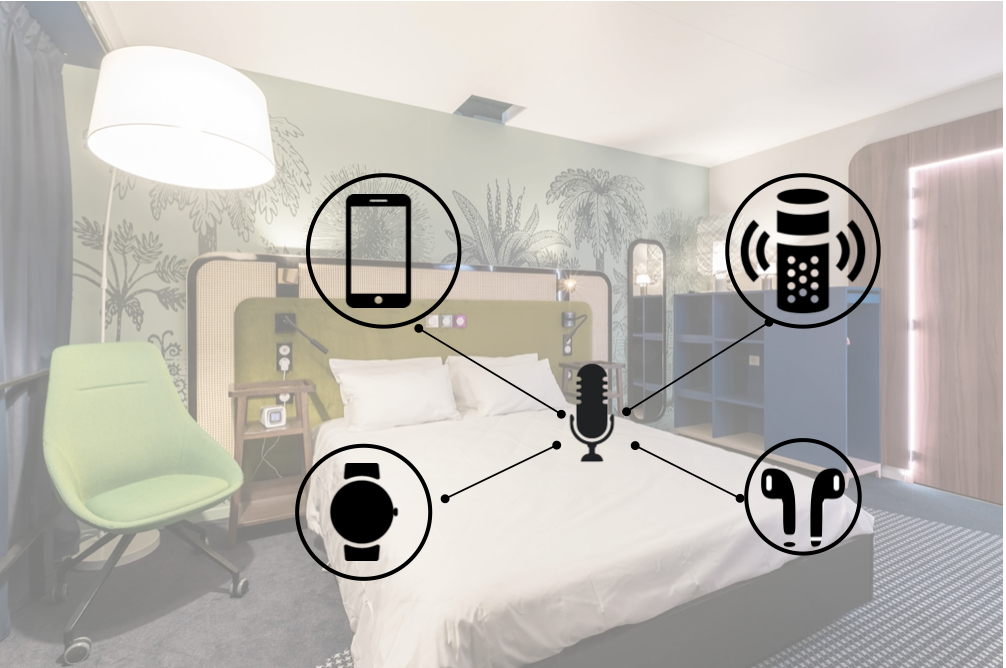}}\hfill
\subfloat[Cameras at a traffic junction]{\label{tsmds-c}\includegraphics[width=0.30\linewidth]{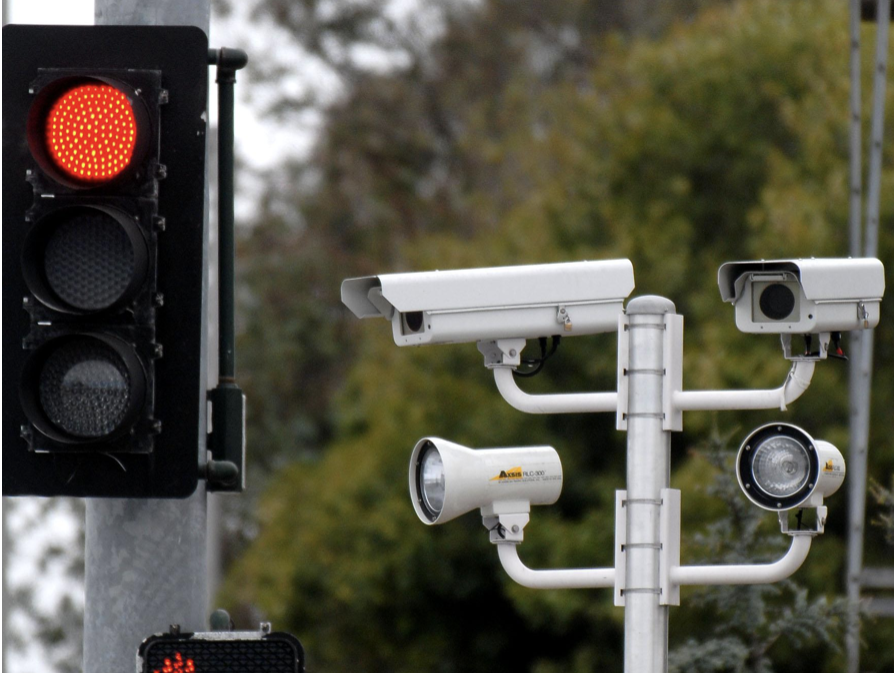}}\par
\caption{Examples of Time-Synchronous Multi-Device Systems (\problem{}) in different contexts.}
\label{fig:tsmds}
\vspace{-0.5cm}
\end{figure}

In this work, we extend the line of research on self-supervised learning for HAR, albeit in a different context. We study a problem setting called Time-Synchronous Multi-Device System~(\problem{}) which opens up an unexplored opportunity for self-supervised learning. This problem setting is inspired by the current trend of people wearing \emph{multiple} intertial measurement unit (IMU)-enabled devices, including smartphones and consumer wearables. Studies (e.g., \cite{deviceby2025}) even estimate that by the year 2025, each person will own 9.3 connected devices on average. An example of this trend is shown in Figure~\ref{tsmds-a} -- here, a user is wearing multiple IMU-enabled devices which are collecting time-synchronized sensor data while the user is performing a physical activity, such as jogging.

Apart from the growing importance and practicality of this problem setting, it presents a unique opportunity for Contrastive Learning for HAR, one of the promising approaches for self-supervised learning. \revisions{Contrastive learning involves comparing a data sample against its \emph{transformed version} and other samples in the dataset. Here, in the \problem{} setting, multiple devices on a user's body are capturing the same physical phenomenon (i.e., a user's activity) from different perspectives. For example, when a user is running, both a smartphone placed in the \emph{thigh pocket} and a smartwatch worn on the \emph{wrist} record sensor data for the \emph{running activity}, but from different perspectives due to their unique placements on the body. Thus, rather than manually generating transformations of the data samples for contrastive learning, we can interpret the data from different devices in the \problem{} setting as natural transformations of each other, and leverage it to design self-supervised contrastive learning algorithms. Here, different devices collaborate in the process of self-supervised learning; hence we call this approach \textbf{Collaborative Self-Supervised Learning} (ColloSSL).}

In \S\ref{sec:prelim}, we provide a formal definition of the \problem{} setting and give a primer on contrastive self-supervised learning. In \S\ref{subsec:motivation}, we elaborate on the limitations of current contrastive learning techniques, which serve as the motivation for our research problem. In \S\ref{sec:solution}, we introduce novel algorithms and optimization objectives for Collaborative Self-Supervised Learning, namely Device Selection, Contrastive Sampling, and a Multi-view Contrastive Loss function. Later, we present an end-to-end semi-supervised framework which uses Collaborative Self-supervised Learning on unlabeled sensor data from multiple devices to learn a high-quality features from the data, which is followed by training an HAR classifier on a small amount of labeled data to recognize specific human activities. Finally, in \S\ref{sec:eval}, we compare the performance of our framework against a number of supervised and semi-supervised training baselines on three multi-device HAR datasets. Our results show that \solution{} generally outperforms supervised training and other semi-supervised baselines in both low-data and high-data regimes. We also present insights on the feature embeddings  generated by \solution{} as well as the robustness and generalizability of \solution{}. 

\parjump{}
\noindent
\revisions{Our work makes three major contributions:}

\begin{itemize}
    \item \revisions{We present Collaborative Self-Supervised Learning (\solution{}), a novel method for learning from unlabeled inertial sensor data collected from multiple devices worn by the user. Different from prior methods~\cite{saeed2019multi, tang2021selfhar}, \solution{} leverages natural transformations in the sensor datasets collected from multiple devices to perform contrastive learning, and learns a robust feature extractor for downstream HAR classification tasks.}
    \item \revisions{We introduce three key research challenges for \solution{} and propose novel device selection and data sampling algorithms, along with a new loss function which extends contrastive learning to a multi-device setting.} 
    \item \revisions{We present a thorough evaluation of \solution{} on three multi-device HAR datasets covering both locomotion activities and complex activities of daily living. Our results show that \solution{} outperforms both fully-supervised and semi-supervised baselines both in terms of recognition accuracy and labeled-data-efficiency.}
\end{itemize}

\section{Related Work}

\revisions{In this section, we review four sets of prior works that are relevant to this paper. Specifically, we review the literature on human-activity recognition (HAR) and the emerging semi-supervised learning techniques aimed at solving the labeled data scarcity challenge in HAR. Closely tied to our work is the recent trend of investigating self-supervised learning in HAR -- as such, we survey the recent ML literature on self-supervised and contrastive learning and explain how it has been applied in HAR. Finally, we also highlight the relation of our work to prior research in multi-device sensing environments. } 

\parjump{}
\noindent
\textbf{Deep Learning for Human Activity Recognition.}
Human Activity Recognition is a classification task, where the labels are activities like ``walking'' and ``running'' etc. that arise naturally in day-to-day activities. The problem has been deeply studied by both signal processing researchers \cite{hammerla2016deep, Guan_2017,10.5555/2832747.2832806,figo2010preprocessing, hammerla2013preserving, plotz2011feature}, utilizing time-series data from sensors embedded in mobile devies, as well as computer vision researchers, who focused on video and camera-based solutions. For any deep learning or machine learning solution in general, the choice of feature extractor plays a crucial role. Traditionally, the solutions were focused on statistical features that have been studied by many researchers \cite{figo2010preprocessing, hammerla2013preserving, plotz2011feature}. Later, deep convolutional and recurrent networks \cite{hammerla2016deep, Guan_2017,10.5555/2832747.2832806} were shown to be effective in learning feature extractors from labeled data. 

\revisions{However, the size, diversity and real-world representativeness of the datasets are crucial for satisfactory real-world performances of machine learning and deep learning solutions~\cite{deep_bias, dataset_bias_1, dataset_bias_2}. Compared to other data modalities such as images, videos and audio clips, where post-hoc annotations are often feasible, obtaining labels for sensor time-series, especially those collected in free-living environments is very difficult~\cite{bulling2014tutorial}. This is exacerbated by the variations between different use-cases and setups, which include different models or types of embedded sensors, placements of sensors, and target activity classes, making it almost infeasible to collect data which can represent all different possible setups. This leads to the paucity and limited quality of labeled data for supervision, which might have an adverse effect on performance in the real world}. 

\revisions{Unsupervised and semi-supervised learning, which include methods that aim to make use of unlabeled data to overcome the limitations of purely-supervised methods, are some of the most popular directions of research in HAR~\cite{semi_supervised_survey, semi_supervised_survey2, semi_har}. Our work also builds upon this line of research on semi-supervised learning techniques, albeit in a different problem setting where we use unlabeled data obtained from multiple devices worn by a user to generate a supervisory signal for representation learning.}

\parjump{}
\noindent
\textbf{Self-supervised Learning for Human Activity Recognition.}
\revisions{Self-supervised learning (SSL) has become an increasingly popular area of research in the machine learning community to minimize the reliance on labeled data for training deep learning models~\cite{khosla2021supervised,simclr_chen2020simple,he2020momentum, caron2021emerging, harley2020learning}. In this vein, a number of self-supervised learning frameworks have been proposed with their unique optimization objectives~\cite{simclr_chen2020simple, grill2020bootstrap, chen2021exploring}. For example, the SimCLR~\cite{simclr_chen2020simple} framework trains a feature extractor to be agnostic against transformations, by using transformed views of the same sample as positive pairs and contrasting them against other samples.}

\revisions{Researchers in the HAR community have recently started exploring how SSL techniques can be either extended or designed specifically for HAR tasks \cite{saeed2019multi,saeed2020sense,tang2020exploring}. In one of the early pioneering works, Saeed et al. used the task of identifying which signal transformation has been applied to a particular data sample as a self-supervision task \cite{saeed2019multi}}. A set of eight signal transformations were chosen to represent signal noises, and the authors reported a performance gain by pre-training the model with this task. However, the authors focused on pre-training the models with data from similar distributions and from a single device. The signal transformation task was also adapted to train models for emotion recognition from electrocardiogram (ECG) data \cite{ecg_transformation_sarkar2020self}. The overall approach follows closely to that of the previous work \cite{saeed2019multi}, with additional explorations on the effects on performance when using pre-training tasks of different levels of difficulties, and the relationship between the tasks used in pre-training and in application. Recently, the SimCLR framework has been applied in HAR \cite{tang2020exploring}. The authors explored a set of different combinations of transformation functions that are designed for time-series data, for training feature extractors for sensor data based on the SimCLR framework. However, this work again focused on leveraging data from a single sensor only, and the potential for extracting stronger supervisory signals from other sensors and devices was not explored. An initial attempt to leverage multiple devices for SSL has been made for visual representation~\cite{tcn_sermanet2018time}. It showed that time-synchronized visual representations can be used to provide a reward function for robot manipulation via reinforcement learning. One of its limitations is that it utilizes data from two camera views only, but our proposal is compatible with settings with more than two data sources, and we thoroughly explored settings with different numbers of IMU devices.

\parjump{}
\noindent
\textbf{Multi-device Environments for HAR:} Multi-device environments for HAR have been actively studied for the past couple decades from various angles. We briefly review the past research on multi-device environments, focusing on sensor selection and fusion for HAR. Firstly, sensor selection strategies have been proposed to maximize the system utility in body sensor networks, by dynamically selecting the best sensor based on predefined parameters of each device, such as average accuracy, resource usage, and device availability~\cite{kang2010orchestrator,keally2011pbn,zappi2008activity,min2019closer}. While these strategies focus on providing better runtime system performance in a multi-device environment, they do not use the data from multiple devices to improve the accuracy of activity recognition. For this purpose, a number of deep learning based sensor fusion techniques~\cite{ordonez2016deep,peng2018aroma,vaizman2018context,yao2017deepsense,yao2018qualitydeepsense} have been proposed, which train a fusion model by concatenating multiple sensor streams. \revisions{While they show significant improvement in model accuracy, they assume that all sensor data streams are labeled. Instead, our work focuses on leveraging \emph{unlabeled} data from multiple devices to learn good quality features from the data, which can later be used to train a downstream HAR model with a very small amount of labeled data. Moreover, in contrast to fusion-based approaches, we do not require the availability of all devices at inference time; the data from multiple devices is only needed during training, and inference can happen independently on each device}.

\section{Preliminaries}
\label{sec:prelim}
In this section, we explain the \problem{} problem setting and provide a short primer on contrastive self-supervised learning.

\subsection{Time-Synchronous Multi-Device Systems}
\label{subsec:tsmds}
\revisions{We begin by providing more details on the problem setting explored in this work called Time-Synchronous Multi-Device Systems (\problem{}). It is important to note that our objective is not to claim that it is a problem setting that has not been studied before; instead, we argue that this is an interesting problem space in which self-supervised learning has not been studied. In this section, we conceptually describe this problem setting and later in \S\ref{subsec.tsmds} we formalize it mathematically.}

\revisions{In the context of human-activity recognition, the \problem{} problem setting is similar to a Body-Area Network in which multiple computing/sensor devices are worn on, affixed to or implanted in a person’s body~\cite{hasan2019comprehensive}. The essential characteristic of \problem{} is that all devices observe a physical phenomenon (e.g., a user's locomotion activity) \emph{simultaneously} and record sensor data in a \emph{time-aligned manner} (Figure~\ref{tsmds-a}). Even though our work focuses on HAR using motion data, the \problem{} setting is rather generic and can be found in many other sensory applications. In a smart home (Figure~\ref{tsmds-b}), multiple voice assistants (e.g., Siri in a smartphone, Alexa and Google Home in a living room) can listen to a user's speech and audio commands simultaneously. In a camera network deployed at a traffic junction (Figure~\ref{tsmds-c}), multiple cameras capture the same scene from different perspectives simultaneously.}  

\parjump
\parjump
\parjump
\parjump
\parjump
\parjump
\parjump
\noindent
Below we state the two key assumptions we have made for exploring \problem{} setting in the context of HAR:

\begin{itemize}
    \item We assume that all sensor devices in the multi-device system share the same sensor sampling rate, or that their data can be re-sampled to the same rate. This assumption ensures that the dimensions of the data that will be fed to the HAR model remain consistent across different devices, and it simplifies the design of the neural network architecture of the HAR model.   
    \item By definition, we assume that multiple devices in the \problem{} setting are collecting sensor data in a time-aligned manner. Admittedly, the assumption that the sensor datasets across multiple devices are perfectly time-aligned is strong in real-world applications. There could be timestamp noise or system clock misalignment across devices, which could skew the temporal alignment of multi-device datasets. However, prior research~\cite{stisen2015smart} in HAR has shown timestamp noise for accelerometer and gyroscope sensors is very small, usually less than 10ms. We hypothesize that such a small noise will not degrade the performance of our solution, and empirically validate this hypothesis by showing that our approach is robust against moderate amounts of temporal misalignment between devices (\S\ref{subsec:misalignment}). 
\end{itemize}

\subsection{Primer on Contrastive Self-Supervised Learning}

Contrastive Learning is a type of self-supervised learning algorithm where the goal is to group similar samples in the dataset closer and dissimilar samples far from each other in the embedding space~\cite{bachman2019learning,simclr_chen2020simple}. In contrastive learning with unlabelled data, a data sample (called the \emph{anchor sample}) from the training dataset is taken and often a pre-defined perturbation (e.g., rotation) is applied to generate a transformed version of the sample. During the training process, the transformed sample is considered as the \emph{positive sample} and other randomly selected data samples from the dataset are considered as \emph{negative samples}. All three types of data samples (anchor, positive, negative) are fed to a neural network to obtain feature embeddings. The goal of the model training is to bring the feature embeddings of the anchor and positive sample closer, while pushing the embeddings of anchor and negative samples far from each other. In this process, the model learns to extract good quality feature embeddings just using unlabeled raw data, and these embeddings could be later used for downstream classification tasks.

One of the most important factors that underpins the performance of contrastive learning is the choice of the perturbation(s) that are applied to each \emph{anchor sample} for which we want the model to remain invariant, while remaining discriminative to other negative samples. Prior research~\cite{simclr_chen2020simple, tang2020exploring} has shown that the choice of perturbations chosen by the algorithm designer can have a profound impact on the performance of contrastive learning. They found that the top-1 accuracy of a model pre-trained with contrastive learning on the ImageNet dataset can drop dramatically from 56.3\% (using an image crop and coloring transformation) to just 2.6\% (using image rotation transformation).

\section{Motivation and Challenges for Collaborative Self-Supervised Learning}
\label{subsec:motivation}
Instead of specifying \emph{manual transformations} (e.g., rotation) during contrastive learning, we ask whether it is possible to leverage \emph{natural transformations} that are already available in sensor datasets. Interestingly, the \problem{} setting presents a unique scenario where we have such natural transformations of HAR data across different devices. As shown in Figure~\ref{tsmds-a}, multiple devices worn by the user are simultaneously capturing the \emph{same physical activity} (e.g., running) from different views. As such, we can consider the dataset from different devices as natural transformations of each other. This observation can also be validated from an early seminal HAR work by Kunze and Lukowicz~\cite{10.1145/1409635.1409639}, where they argued that the accelerometer and gyroscope data collected by body-worn sensors depend on the translation and rotation characteristics of the body part where the sensor is placed. Since different body parts have different translation and rotation characteristics (e.g., wrist has a higher rotational degree of freedom than chest), it induces natural transformations in the accelerometer and gyroscope data collected from different body positions. 

In summary, the \problem{} setting allows us to capture time-aligned sensor datasets which have natural transformations across them. Our idea is to leverage these natural transformations to define a pretext task and perform contrastive learning. 

\begin{figure}
    \centering
    \includegraphics[width=0.8\linewidth]{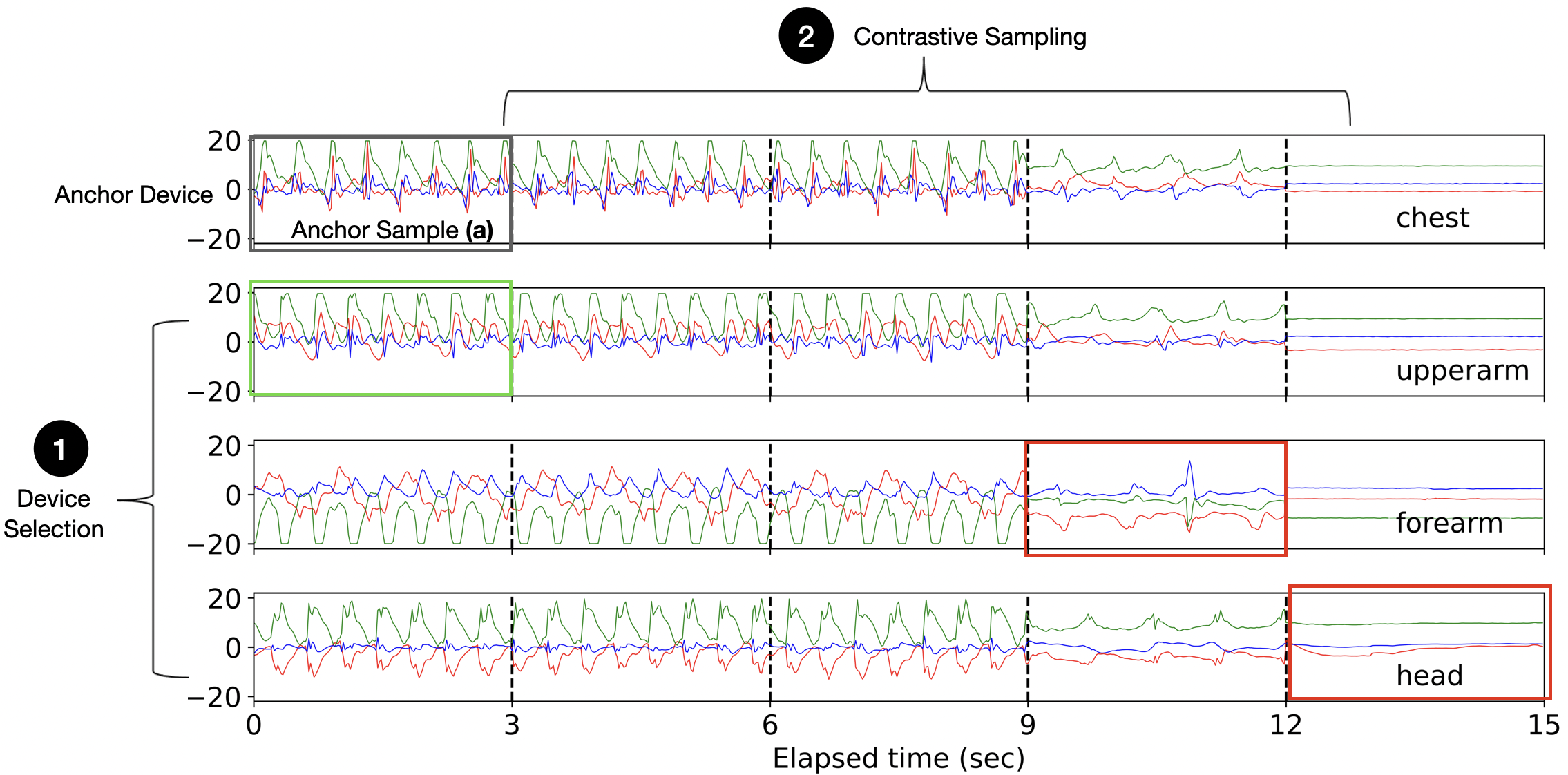}
    \vspace{-0.2cm}
    \caption{\revisions{Illustration of two research challenges in \solution{}: Device Selection and Contrastive Sampling. The anchor sample is denoted by the grey rectangle. The green and red rectangles denote the positive and negative samples that are selected by our Device Selection and Contrastive Sampling algorithms described in \S\ref{sec.dds-cs}.}}
    \vspace{-0.4cm}
    \label{fig:gsl_motivation}
\end{figure}

\parjump
\noindent
\textbf{Research Challenges in \solution{}.} \revisions{To motivate our research challenges, we present an example illustration in Figure~\ref{fig:gsl_motivation}. The figure shows a 15-second trace of unlabeled accelerometer data collected simultaneously from $N$ (=4) body-worn devices. Each accelerometer trace is segmented using a window length of 3 seconds, thereby giving us 5 windows/samples for each device. Let us say we would like to train a feature extractor for the `chest' body position using contrastive learning. As such, the data samples from `chest' would become the \emph{anchor samples}. The first anchor sample from `chest' \textbf{($a$)} is highlighted by a grey rectangle in Figure~\ref{fig:gsl_motivation}. As explained above, to perform contrastive learning, we need to select appropriate positive and negative samples. Below we identify three key research questions in this direction:}

\begin{itemize}
    \item From the remaining $N-1$ devices (i.e., upperarm, forearm, head), how do we select positive and negative samples for contrastive training? Intuitively, there will be some devices whose data distribution will be too far apart from the data distribution of `chest'. If we use these far-apart devices to obtain \emph{positive samples} and push them closer in the embedding space to the anchor samples, it may lead to degenerate solutions for contrastive learning. Further, as the data distribution of each device changes depending on the user's activity, we need to account for these dynamic changes while selecting devices. \revisions{We call this research challenge as \emph{Device Selection} \circled{1} and propose a data-driven strategy which uses the Maximum Mean Discrepancy (MMD) as a metric to dynamically select positive and negative devices during contrastive learning. As an illustration, we show in Figure~\ref{fig:gsl_motivation} that our strategy chooses `upperarm' as the positive device and `forearm' and `head' as the negative devices for the given data samples. The selection algorithm is described in detail in \S\ref{subsec:device_selection}.} 
    
    \item In addition to \emph{Device Selection}, we need to answer which samples from the selected devices will act as positive or negative samples. For example, if we select `upperarm' as the positive device, which one of its 5 samples (or {windows}) will act as the positive sample for the anchor sample \textbf{($a$)}? \revisions{We call this challenge \emph{Contrastive Sampling} \circled{2} and propose the idea of using time-synchronized samples from positive devices and time-asynchronized samples from negative devices. For example, in Figure~\ref{fig:gsl_motivation}, the time-synchronized sample from the positive device is highlighted with a green rectangle, whereas the time-asynchronized samples from negative devices are highlighted with red rectangles. The rationale and details behind the contrastive sampling algorithm are provided in \S\ref{subsec:contrastive_sampling}.}
    
    \item Finally, to enable contrastive learning in a group of devices, we need to define a new optimization objective. To this end, we propose a novel loss function called \emph{Multi-View Contrastive Loss} which can take an arbitrary number of positive and negative samples as input and compute a loss metric, which is then optimized using stochastic gradient descent (\S\ref{subsec.gcl}). 

\end{itemize}

\section{Collaborative Self-Supervised Learning}
\label{sec:solution}

In this section, we introduce our proposed approach of Collaborative Self-Supervised Learning for HAR. 

\begin{figure}
    \centering
    \includegraphics[width=0.9\linewidth]{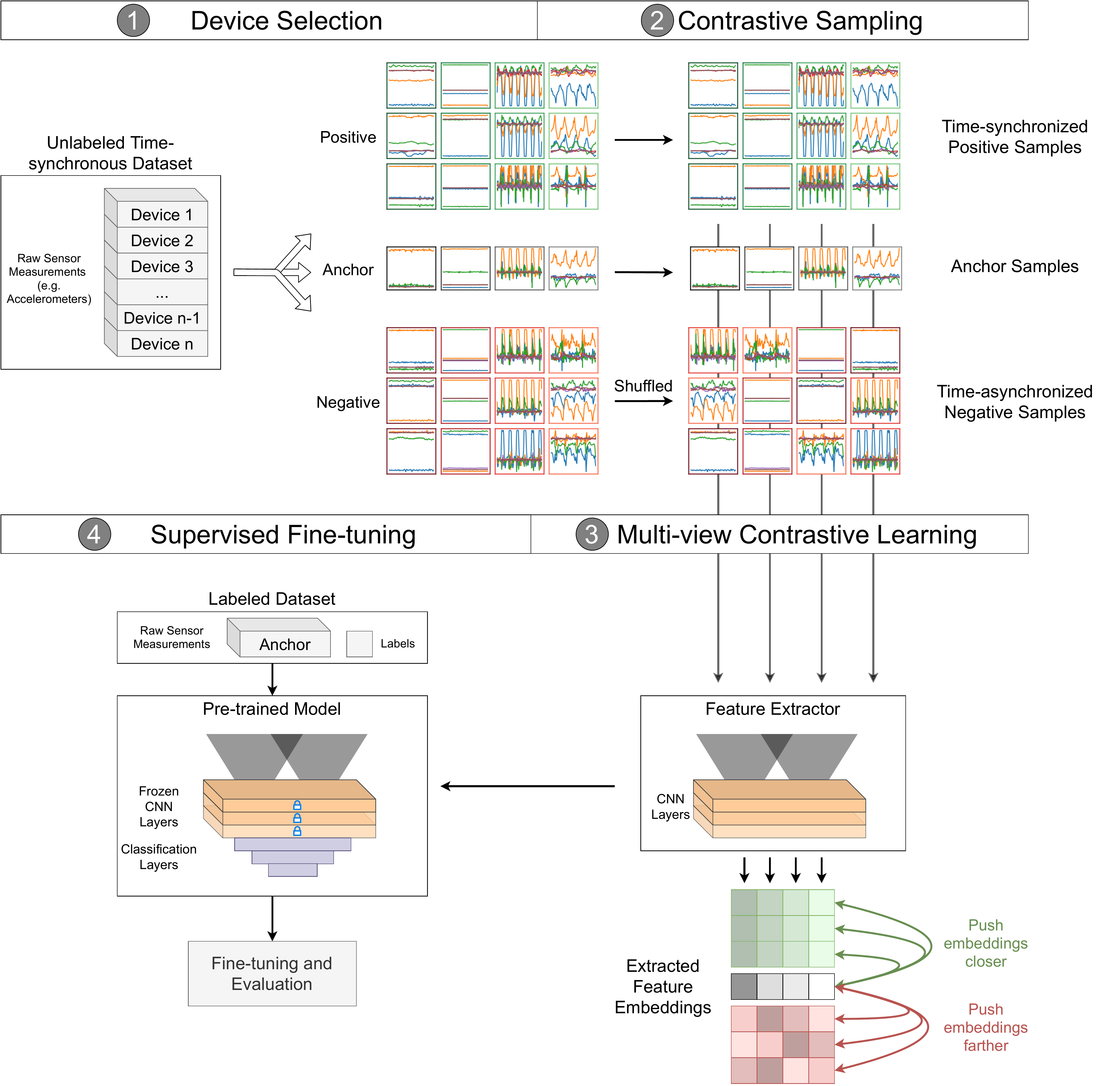}
    \caption{Overview of Collaborative Self-Supervised Learning. Please refer to \S\ref{subsec:sol_overview} for a detailed explanation.}
    \label{fig:gsl_architecture}
\end{figure}

\subsection{Problem Statement}
\label{subsec.tsmds}

We formalize the \problem{} problem setting as follows: we are given $D$ devices with time-aligned and unlabeled sensor datasets $\{\mathbb{X}_i\}_{i=1}^{D}$. Without loss of generality, we assume that the datasets are pre-segmented into $T$ windows, as is the convention in HAR tasks. Each dataset $\mathbb{X}_i$ contains $T$ windows $\{x_i^1,\cdots,x_i^T\}$, where $x_i^j$ denotes a set of sensor samples from device $i$ in window $j$. In general, a sensor sample could be any form of IMU measurement, e.g., 3 dimensional accelerometer data, 3 dimensional gyroscope data, 1 dimensional accelerometer norm. In our work, each sensor sample is a 6-dimensional vector, created by stacking together 3-axis of accelerometer and 3-axis of gyroscope data. 

Let $D^* \in D$ be an anchor device (e.g., a smartphone) for which we want to train an HAR prediction model. Let $\mathbb{L}^* = \{(x_*^1, y_*^1)\cdots,(x_*^m, y_*^m)\}$  be a (small) pre-segmented labeled dataset from the anchor device with $m$ windows ($m \ll T$) where $x_*^j$ is the set of sensor samples in window $j$ and $y_*^j$ is the class label assigned to those samples.       

\parjump{}
\noindent
Our objectives are two-fold: first, we aim to use the unlabeled datasets $\{\mathbb{X}_i\}_{i=1}^{D}$ from all the devices to train a feature extractor $f(.)$ using \solution{}. The trained feature extractor should be able to generate high-quality feature embeddings for the anchor device data.
Second, we aim to use this pre-trained feature extractor $f(.)$ to obtain feature embeddings for the labeled anchor samples $x_*^j$, and subsequently train an HAR classifier $g(.)$ which maps these features embeddings to the corresponding class labels $y_*^j$. 

\subsection{Solution Overview}
\label{subsec:sol_overview}

Our proposed solution is illustrated in Figure~\ref{fig:gsl_architecture} and works as follows:

\begin{enumerate}[label={\arabic*.}]
    \item We initialize the feature extractor $f(.)$ with random weights. 
    \item We sample a batch $B$ of time-aligned unlabeled data $\{x_i^1,\cdots,x_i^B\}_{i=1}^{D}$ from all $D$ devices. 
    \item \textbf{Device selection} \grey{1} in Figure~\ref{fig:gsl_architecture}: Given an anchor device $D^*$, we first decide which of the remaining devices will provide \emph{positive samples} and \emph{negative samples} for contrastive learning in this batch of data. \revisions{This is done through a \emph{Device Selection} algorithm explained in \S\ref{subsec:device_selection}.} 
    \item \textbf{Contrastive sampling} \grey{2} in Figure~\ref{fig:gsl_architecture}: Next, for each anchor sample, we decide out of all samples in the batch $B$, which specific samples from the positive and negative devices should be used for contrastive learning. \revisions{While data sampling has been studied as an important aspect of SSL in the past, we present an algorithm called \emph{Contrastive Sampling}  (\S\ref{subsec:contrastive_sampling}) which extends data sampling to the \problem{} setting.} 
    \item  \textbf{Multi-view contrastive learning} \grey{3} in Figure~\ref{fig:gsl_architecture}: The anchor, positive and negative sample(s) are fed to the feature extractor $f(.)$ to generate feature embeddings. These feature embeddings are used to compute a Multi-view Contrastive Loss (detailed in \S\ref{subsec.gcl}), which pushes the positive embeddings closer to the anchor, and negative embeddings far from the anchor.
    \item Steps 4-5 are repeated until all anchor samples in the batch $B$ are computed, and Steps 2-5 are repeated until the Multi-view Contrastive Loss converges. Upon convergence, we expect that $f(.)$ has learned to extract high-quality features for the anchor device. 
    \item \textbf{Supervised fine-tuning} \grey{4} in Figure~\ref{fig:gsl_architecture}: Finally, we use the pre-trained feature extractor $f(.)$ and the labeled dataset from the anchor device ($\mathbb{L}^*$) to train an HAR classifier using supervised learning (\S\ref{subsec.finetune}). 

\end{enumerate}

\subsection{Feature Extractor}
\label{subsec.feature_extractor}

We use a temporal (1-dimensional) convolutional neural network as the feature extractor. This design choice is inspired by prior work on self-supervised learning in HAR~\cite{saeed2019multi} and allows us to do a fair comparison of our solution against the baselines by keeping the model architecture consistent across different techniques. The architecture consists of three 1D convolutional layers, with 32, 64 and 96 filters and a kernel size of 24, 16, and 8 respectively, all with stride 1. For regularization, we use Dropout between every pair of layers with a rate of 0.1, and apply L2 regularization with a regularization factor of 1e-4. Finally, the output of the last Conv1D layer is passed to a GlobalMaxPooling (1D) layer. 

The input to $f(.)$ is a 2-dimensional tensor, with time on one axis and sensor data on the other. As we use both 3-axis accelerometer and 3-axis gyroscope traces as the input, the dimension of sensor data is 6. As such, given a sampling rate of 50 Hz and a window length of 2 seconds, the input tensor to the model is of dimension $100 \times 6$. The output of $f(.)$ is a feature embedding with dimension $(96 \times 1)$. 

\subsection{Device Selection and Contrastive Sampling}
\label{sec.dds-cs}

As explained in \S\ref{subsec:motivation} and Figure~\ref{fig:gsl_motivation}, two key challenges in \solution{} are: (1) \emph{Device Selection}, i.e., selecting the devices from which positive and negative samples will be taken, and (2) \emph{Contrastive Sampling}: deciding which samples from the selected devices will be used for contrastive learning. 

\parjump{}
\noindent
Before we present our approach, we first explain that for a given anchor sample ($x$), what makes a `good' positive ($x^+$) and negative ($x^-$) sample for contrastive learning. Recall that the objective of contrastive learning is to guide the feature extractor $f$ to map $x$ and $x^+$ closer to each other in the feature space, and $x$ and $x^-$ far from each other. 

\parjump{}
\noindent
\textbf{Goodness of positive samples.} We propose that a good positive sample $x^+$ should have the following two characteristics: 

\begin{enumerate}[label=\textbf{(P\arabic*)}]
    \item {$x^+$ should belong to the same label/class as the anchor sample $x$}. Because if $x$ and $x^+$ are from different classes and yet the feature extractor $f$ tries to map them closer to each other, it would lead to poor class separation and degrade the downstream classification performance. 
    
    \item {$x^+$ should come from a device whose data distribution has a small divergence from that of the anchor device.} This is important because if the anchor device and positive device are very different in their data characteristics (e.g., wrist-worn IMU vs. chest-worn IMU), then it might affect the feature extractor in extracting meaningful invariant embeddings.
\end{enumerate}

Note that in \solution{}, we do not have access to the ground-truth labels of the data. As such, enforcing \textbf{(P1)} on $x^+$ may seem tricky at first. However, from the definition of \problem{} setting, we know that all devices collect the HAR data simultaneously and in a time-aligned fashion. Hence, we can naturally assume that the ground-truth labels (e.g., walking, running) are also time-aligned across devices. Therefore, if we can ensure that $x$ and $x^+$ are time-aligned, it will implicitly guarantee that they have the same labels. 

\parjump{}
\noindent
\textbf{Goodness of negative samples.} We propose that a good negative sample $x^-$ should have the following two characteristics:

\begin{enumerate}[label=\textbf{(N\arabic*)}]
    \item {$x^-$ should be a true negative, i.e., belong to a different class than the anchor sample $x$}. Because if $x$ and $x^-$ are from the same class and yet the feature extractor $f$ tries to push them away, it could lead to poor classification performance.  
    
    \item The most informative negative samples are those whose embeddings are initially near to the anchor embeddings, and $f$ needs to push them far apart. In this scenario, $f$ gets a strong supervisory signal from the data and more useful gradients during training. In the alternate scenario when negative embeddings are already far apart from anchor when the training initializes, $f$ will receive a weaker supervisory signal from the data, which could adversely impact its convergence. 
    
\end{enumerate}

Again, due to the unavailability of class labels in \solution{}, strictly enforcing \textbf{(N1)} is not possible. A solution is needed which can encourage this characteristic on negative samples, and minimize the possibility of $x$ and $x^-$ belonging to the same class. 

\parjump{}
\noindent
Having defined what constitutes a good positive and negative sample, we now describe our device selection and contrastive sampling algorithms. 

\subsubsection{Device Selection}
\label{subsec:device_selection}
 Our device selection algorithm is designed to increase the likelihood of selecting `good' positive and negative samples. For brevity, we refer to the devices from which positive (or negative) samples are taken as \emph{positive devices} (or \emph{negative devices}). Formally,  we are given a set of $D$ devices with time-aligned and unlabeled sensor datasets $\{\mathbb{X}_i\}_{i=1}^{D}$. Let $D^* \in D$ be the anchor device. Let $D^{\theta} = {D}\setminus{D^*}$ be a \emph{candidate set} of remaining {devices} from which we want to choose positive and negative devices. 

\parjump{}
\noindent
Our device selection algorithm works as follows: first, we sample a batch of time-aligned data from the anchor device $D^*$ and each of the device in $D^{\theta}$. Let $x^*$ and $X^\theta = \{x_i\}_{i=1}^{|D^{\theta}|}$ be the data batches from the anchor and the candidate devices, respectively. 

\parjump{}
\noindent
We compute the pairwise Maximum Mean Discrepancy (MMD) between $x^*$ and each of the data batches in $X^\theta$. MMD is a distribution-level statistic to compute the distance between two data distributions; a higher MMD implies a larger distance between distributions~\cite{gretton2008kernel}. After obtaining the batch-wise MMD scores between each pair of device batches, we use the following device selection policy:

\parjump{}
\noindent
\textbf{Closest Positive}. The device whose data has the least MMD distance from the anchor data is chosen as the positive device. This choice satisfies the criteria \textbf{(P2)} for selecting good positive samples and ensures that $f(.)$ can reasonably map the embeddings of the two samples closer to each other. Note that we also experimented with using more than one positive device, but found that using just one (closest) device as positive gives the best performance. 

\parjump{}
\noindent
\textbf{Weighted Negatives}. For negative devices, we use `all' devices from the candidate set $D^{\theta}$, but their contributions during training are weighted by the inverse of their MMD distance from the anchor samples. Devices which have smaller MMD distance to anchor get higher weights during contrastive training, and devices with higher MMD distances get smaller weights. This policy serves two objectives: firstly, by assigning higher weights to devices with smaller MMD distances to the anchor, it satisfies \textbf{(N2)} and ensures that those negative samples which are closer to the anchor get more weight during training. Secondly, the use of `all' devices as negatives serve as a hedge against the scenario when \textbf{(N1)} is violated on one device. For example, even if one device violates \textbf{(N1)} and ends up choosing $x^-$ from the same class as $x$, the other devices can cover for it, and ensure that its impact on the feature extractor is minimal. 

The weights assigned to each negative device $i \in D^{\theta}$ can be expressed as:
\begin{equation}
 w_i = \frac{1}{MMD \left( x^*, x_i \right)} 
 \label{eq:wi}
\end{equation}
The weights are further normalized by dividing each weight with the maximum weight across devices. 

\parjump{}
\noindent
As an example, we apply our device selection policy to the RealWorld HAR dataset (details of the dataset are provided in \S\ref{subsec:experiment_setup}). The dataset contains sensor data from 7 IMU-equipped devices: $D =$ \{chest, upperarm, forearm, thigh, shin, head, waist\}. We choose `chest' as the anchor device and obtain the pairwise MMDs between data from `chest' and data from the remaining devices. This results in the following pairwise MMD scores: \{\emph{chest-head: 0.45, chest-waist: 0.61, chest-thigh: 0.67, chest-upperarm: 0.77, chest-forearm: 0.83, chest-shin: 1.51}\}. In line with our selection algorithm, we choose \emph{head} as the positive device and \{\emph{head, waist, thigh, upperarm, forearm, shin}\} as the negative devices with weights inversely proportional to their MMD scores.

\subsubsection{Contrastive Sampling} 
\label{subsec:contrastive_sampling}
At the previous stage, we have decided which devices in $D_\theta$ will act as positive and negative. Now, we decide which data samples should be picked from each device for contrastive training. 

\parjump{}
\noindent
Formally, we are given an anchor device $D^*$, a set of positive ($D^+$) and negative devices ($D^-$). Let $P_i =\{p_i^1,\cdots,p_i^T\}|_{i=1}^{|D^+|}$ , $N_j =\{n_j^1,\cdots,n_j^T\}|_{j=1}^{|D^-|}$, $A =\{a^1,\cdots,a^T\}$ be the time-aligned data batches from the $i^{th}$ positive, $j^{th}$ negative, and the anchor device respectively. Here, $p_i^t$, $n_j^t$ and $a^t$ each denote a data sample at time-step $t$.

\parjump{}
\noindent
The objective of contrastive sampling is to select `good' positive and negative embeddings for a given anchor sample $a^t$. Our sampling policy works as follows:

\parjump{}
\noindent
\textbf{Synchronous Positive Samples.} For a given anchor sample $a^t$ at time-step $t$, we choose its time-aligned positive counterparts $p_i^t$ as the positive samples. As explained earlier, this choice ensures that the anchor and positive samples have the same labels, and satisfies the \textbf{(P1)} criteria for good positive samples.

\parjump{}
\noindent
\textbf{Asynchronous Negative Samples.} A criteria for good negative samples \textbf{(N1)} is that they should belong to a different class from the anchor sample. As we do not have access to ground-truth labels during contrastive learning, it is impossible to strictly enforce \textbf{(N1)}. As a solution, we make use of the observation that negative samples which are \emph{not time-synchronized} with the anchor are more likely to be from a different class. That is, for an anchor sample $a^t$ at time-step $t$, a good negative sample would be $n^{t'} \mid~t' \neq t$. 

This choice however still does not guarantee that the labels at time-steps $t$ and $t'$ will be different; for example, a user's activity at $t=0$ and $t'=100$ may happen to be the same by random chance. To minimize the possibility of such cases, we use a simple trick: a large batch size of 512 is used during \solution{} which ensures that each batch has diverse class labels in it and the possibility of a label overlap at $t$ and $t'$ by random chance is reduced.  

\subsubsection{Summary}
The techniques presented in this section address the two core research challenges of \solution{} identified in Figure \ref{fig:gsl_motivation}. Using the Device Selection algorithm, we first decide which of the devices will act as positive or negative during training. Next, the Contrastive Sampling algorithm finds the `good' positive and negative samples from the selected devices, which can be used for contrastive learning with the anchor embedding. 

A curious reader may have noted that the positive and negative devices selected by our algorithm are not mutually exclusive. The positive device which has the least MMD distance from the anchor will also get selected in the set of negative devices. During Contrastive Sampling, however, the samples selected from this device will differ: when it acts as the positive device, samples which are time-synchronized with the anchor will be selected. However, when it acts as a negative device, samples which are not time-synchronized with the anchor will be selected.

\subsection{Multi-view Contrastive Loss}
\label{subsec.gcl}
In this section, we explain how the positive and negative samples selected from the previous step are used to train the feature extractor. Firstly, the anchor sample, positive samples(s) and negative samples are fed to the feature extractor to obtain feature embeddings. Let $\{z_i^+\} |_{i=1}^{|D^+|}$ and $\{z_j^-\}|_{j=1}^{|D^-|}$ be the selected feature embeddings from the $i^{th}$ positive and $j^{th}$ negative device. Let $z^*$ be the anchor embedding. 

We propose a novel loss function called Multi-view Contrastive Loss, which is inspired by the standard contrastive loss function but compatible with multiple positive and negative samples. 
\begin{align}
    \boldsymbol{\mathcal{L}_{MCL}} =  - \log \frac{\sum_{i=0}^{|D^+|}\text{exp}\left(\text{sim}\left(z^*,z_i^+\right)/\tau\right)} {\left(\splitdfrac{\sum_{i=0}^{|D^+|}\text{exp}\left(\text{sim}\left(z^*,z_i^+\right)/\tau\right)}{ + \sum_{j=0}^{|D^-|}w_j \text{ exp}\left(\text{sim}\left(z^*,z_j^-\right)/\tau\right)}\right)} \label{eq:gsl-2}
\end{align}
where sim(.) denotes \emph{cosine similarity}, $w_j$ is weight assigned to the $j^{th}$ negative device according to \eqref{eq:wi}, and $\tau$ is a hyperparameter denoting temperature. 

\parjump{}
\noindent
$\boldsymbol{\mathcal{L}_{MCL}}$ is minimized for each batch of data using stochastic gradient descent. Effectively, the loss minimization during training guides the feature extractor $f(.)$ to increase the cosine similarity between anchor and positive embeddings (i.e., push them closer in the feature space), and do the opposite for anchor and negative embeddings. In doing so, $f(.)$ understands the structure of the sensor data from different devices, and learns to map raw data into good quality features, which can be useful for various downstream classification tasks. 

\subsubsection{Supervised Fine-tuning} 
\label{subsec.finetune}
Finally, after the feature extractor is trained using \solution{}, it can be used for training down-stream HAR classification models. To this end, we follow the approach by Saeed et al.~\cite{saeed2019multi} of freezing the weights of the feature extractor except its last convolution layer and adding a classification head to the model. The classification head consists of a fully connected layer of 1024 hidden units with ReLU activation, followed by an output layer with the number of units equal to the number of labels. The model is then trained with a small labeled dataset $\mathbb{L}^*$ from the anchor device by optimizing the standard Categorical Cross Entropy loss. 
\section{Evaluation}
\label{sec:eval}
We evaluate \solution{} on three multi-device HAR datasets, and compare its performance against various HAR baselines such as self-supervised learning, semi-supervised learning and fully-supervised learning. Our key results include: 

\begin{itemize}
    \item \solution{} outperforms the fully-supervised learning in a low-data regime. In \revisions{15 out of the 18 anchor devices}, \solution{} trained with 10\% or 25\% of the labeled data achieves higher $F_1$ score than the fully-supervised model trained with 100\% labeled data.
    \item \solution{} also outperforms various HAR baselines in terms of recognition performance. When the same amount of labeled data is used, \solution{} has an absolute increase of \revisions{7.9\% in the $F_1$ score}, compared to the best performing baseline.
    \item Through visualization of t-SNE plots and saliency maps, we show that \solution{} generates well-separable and meaningful feature embeddings across classes. 
    \item \solution{} is robust to temporal misalignment of data from multiple devices; less than \revisions{$\pm$0.006 difference in the $F_1$ score is observed when up to 0.5s} and less than \revisions{$\pm$0.01 difference when up to 3s} of time synchronization error is introduced in devices.
\end{itemize}

\subsection{Experimental Setup}
\label{subsec:experiment_setup}
\begin{table}[t]
	\small
	\begin{tabularx}{\textwidth}{m{0.22\textwidth}|P{0.01\textwidth}X|P{0.1\textwidth}|P{0.01\textwidth}X}
		\hline
		Dataset & \multicolumn{2}{c|}{No. of devices (positions)} & No. of users & \multicolumn{2}{c}{No. of activities (labels)} \\
		\hline \hline
		RealWorld~\cite{sztyler2016body} & 7 & (forearm, thigh, head, upper arm, waist, chest, shin) & 15 & 8 & (stair up, stair down, jumping, lying, standing, sitting, running, walking) \\
		\hline
		Opportunity~\cite{roggen2010collecting} & 5 & (back, left lower arm, right shoe, right upper arm, left shoe) & 4 & 4 & (standing, walking, sitting, lying) \\
		\hline
		PAMAP2-  Locomotion~\cite{reiss2012introducing} & 3 & (arm, chest, ankle) & 8 & 4 & (standing, walking, sitting, lying) \\
		\hline
		PAMAP2-ADL~\cite{reiss2012introducing} & 3 & (arm, chest, ankle) & 8 & 12 & (standing, walking, sitting, lying, running, cycling, nordic walking, ascending stairs, descending stairs, vacuum cleaning, ironing, rope jumping) \\
		\hline
	\end{tabularx}
	\caption {Summary of datasets used for evaluation. \revisions{The \textsc{RealWorld} and \textsc{Opportunity} datasets also contain heterogeneous sensors from different manufacturers.}}
	\label{table:eval_dataset_summary}
    \vspace{-0.4cm}
	
\end{table}

\textbf{Datasets:} For our experiments, we use three datasets for human activity recognition (HAR) which have time-aligned sensor data from multiple devices: \textsc{RealWorld}~\cite{sztyler2016body}, \textsc{Opportunity}~\cite{roggen2010collecting}, and \textsc{PAMAP2}~\cite{reiss2012introducing}
as shown in Table~\ref{table:eval_dataset_summary}. In common, they contain 3-axis accelerometer and 3-axis gyroscope data sampled simultaneously from multiple on-body devices. \revisions{The inertial sensors used in two of the datasets are also heterogeneous: the \textsc{RealWorld} dataset uses a Samsung Galaxy S4 smartphone and a LG G smartwatch to collect the sensor data, while the inertial sensors used in the \textsc{Opportunity} dataset also come from different manufacturers such as InertiaCube3 and Sun SPOT.}

\begin{itemize}
	\item \textbf{RealWorld:} The RealWorld dataset~\cite{sztyler2016body} contains accelerometer and gyroscope traces of 15 participants, sampled at 50 Hz simultaneously on 7 sensor devices mounted at forearm, thigh, head, upper arm, waist, chest, and shin. Each participant performed 8 activities: \emph{climbing stairs down and up, jumping, lying, standing, sitting, running/jogging, and walking}.

	\item \textbf{Opportunity:} The Opportunity dataset~\cite{roggen2010collecting} consists of IMU data collected from 4 participants performing activities of daily living with 17 on-body sensor devices, sampled at 30 Hz. For our evaluation, we select five devices deployed on back, left lower arm, right shoe, right upper arm, and left shoe, and we target to detect the mode of locomotion classes, namely \emph{standing, walking, sitting, and lying}.

	\item \textbf{PAMAP2 \revisions{(Locomotion and ADL)}: } The PAMAP2 Physical Activity Monitoring dataset~\cite{reiss2012introducing} contains data of 18 different physical activities, performed by 9 participants with 3 IMUs. The IMUs were deployed over the wrist on the dominant arm, on the chest, and on the dominant side's ankle with a sampling rate of 100 Hz. \revisions{Out of 9 participants, we used the data from 8 of them, since the remaining user has data for only one activity class. We performed the evaluations using two different splits of the dataset: PAMAP2 - Locomotion, which consists of 4 locomotion activities, \emph{standing, walking, sitting, and lying}, and PAMPA2 - ADL, which consists of 12 ADL activities: \emph{running, cycling, nordic walking, ascending stairs, descending stairs, vacuum cleaning, ironing, and rope jumping}, along with the four locomotion activities.}

\end{itemize}

\noindent
\textbf{Baselines:} We compare \solution{} against 6 baselines, divided in the following 4 categories:

\begin{itemize}
	\item \textbf{Random:} In the \textsc{Random} baseline, we assign random weights to the feature extractor and freeze them. During the supervised fine-tuning, only the classification head is trained using labeled data from the anchor device. This baseline is used to confirm that our learning task is not so trivial that it can be solved with a random feature extractor. 
	\item \textbf{Supervised}: To represent \textit{supervised} learning, \revisions{we devise two baselines, \textsc{Supervised-single} and \textsc{Supervised-multi}. In both baselines, the feature extractor and classifier are jointly trained using labeled data by optimizing a cross entropy loss. In \textsc{Supervised-single}, a separate model is trained on the labeled data of each anchor device. On the other hand, \textsc{Supervised-multi} trains a common model using labeled data from all devices present in the dataset.}

	\item \textbf{Semi-supervised:} As example of semi-supervised learning, we use two AutoEncoder~\cite{pmlr-v27-baldi12a} baselines: \textsc{AutoEncoder-single} and \textsc{AutoEncoder-multi}. In both these baselines, the feature extractor acts as an `encoder' that converts unlabeled input samples into feature embeddings. We add a separate `decoder' neural network which does the inverse task, i.e., it tries to map the feature embeddings back to the input samples. The encoder and decoder together form the AutoEncoder (AE) and are trained by minimizing the MeanSquaredError between the input data and the reconstructed data. In \textsc{AutoEncoder-single}, a separate AE is trained for each anchor device, \revisions{where in \textsc{AutoEncoder-multi}, a common AE is trained using data from all devices.} After the AE training converges, we discard the decoder and use the trained encoder as our feature extractor $f(.)$. Subsequently,  $f(.)$ is fine-tuned on the labeled data from the anchor device. 
	
	\item \textbf{Self-supervised:} To compare the performance of \solution{} with a state-of-the-art self-supervised learning (SSL) technique, we adopt the \textsc{Multi-task SSL} technique proposed by Saeed et al.~\cite{saeed2019multi}. Multi-task SSL learns a multi-task temporal convolutional network with unlabeled data for transformation recognition as a pretext task \revisions{to learn a general feature extractor} and \revisions{trains a} recognition model \revisions{based on the feature extractor} with a small labeled dataset. \revisions{The fine-tuning stage aligns the model to the data distribution of a particular device, and to ensure fair comparisons across different baselines on specialized models, separate recognition models are trained for each anchor device.} Please note that although there are other SSL techniques proposed for HAR, we chose ~\cite{saeed2019multi} as a baseline, because it also applies transformations to the sensor data values, thus making it a fairer comparison against \solution{}. 
	
\end{itemize}

\noindent
\textbf{Data pre-processing and hyperparameters:} The accelerometer and gyroscope traces are segmented into time windows of 3 seconds for RealWorld and 2 seconds for Opportunity and PAMAP2 datasets without any overlap. These window sizes are chosen based on prior explorations with these datasets, e.g., \cite{chang2020systematic, liono2016optimal}. Finally, the dataset was normalized to be in the range of -1 and 1.

Our training setup is implemented in Tensorflow 2.0. We used the TF HParams API\footnote{\url{https://www.tensorflow.org/tensorboard/hyperparameter_tuning_with_hparams}} for hyperparameter tuning and arrived at the following training hyperparameters: \{\solution{} learning rate: 1e-5, fine-tuning and supervised learning rate: 1e-3, $\tau = 0.05$, batch size = 512 \}.    

\parjump
\noindent
\textbf{Evaluation Protocol and Metrics:} In \solution{}, even though we use unlabeled data from multiple devices to train the feature extractor, our evaluation is always done on a single anchor device (e.g., chest-worn IMU for \textsc{RealWorld}). We divide the participants into multiple groups and conduct leave-one-group-out cross-validation. The number of groups of RealWorld, Opportunity, and PAMAP2 is set to 5, 4, and 4, respectively. More specifically, we train the feature extractor using unlabeled data from all groups except a \emph{held-out} group. Thereafter, the feature extractor along with the classification head is fine-tuned on a labeled dataset from the anchor device. Finally, the fine-tuned model is tested on the data from the \emph{held-out} group. 

We use the macro $F_1$ score (unweighted mean of $F_1$ scores over all classes) as a performance metric, which is considered a reasonable evaluation strategy for imbalanced datasets~\cite{plotz2021applying}. The weighted $F_1$ score is also known to take class imbalances into account, however as argued by Pl{\"o}tz \cite{plotz2021applying}, it could inflate recognition results in the favor of majority class.

\subsection{Performance of \solution{} in a Low-data Regime} \label{subsec:performance_low_data}
We evaluate the HAR performance of \solution{} against baseline techniques in two aspects. First, we study whether our solution performs on-par compared to the baselines in a low-data regime, i.e., whether \solution{} shows comparable performance even with a small amount of labeled data. Second, we investigate whether our solution outperforms the baselines in terms of the recognition accuracy, with the same data availability, i.e., when all baselines are trained with the same amount of labeled data.

\begin{table}[t]
	\small
	\begin{tabularx}{\textwidth}{p{0.1\textwidth}|p{0.1\textwidth}|p{0.11\textwidth}p{0.11\textwidth}p{0.11\textwidth}p{0.11\textwidth}p{0.11\textwidth}p{0.11\textwidth}}	
		\hline
		\specialcell{Dataset\\(anchor)} & \specialcell{Supervised-\\single} & Random & \specialcell{Supervised-\\multi} & \specialcell{AutoEncoder-\\single} & \specialcell{AutoEncoder-\\multi} & \specialcell{Multi-task\\SSL} & \solution{} \\
		\hline \hline
		\multicolumn{8}{l}{\textbf{RealWorld}} \\
		\hline
forearm           & 0.732 (100\%) & 0.253 (50\%)*  & 0.495 (100\%)*     & 0.723 (100\%)*  & 0.739 (75\%)   & 0.734 (50\%)   & \textbf{0.767} \red{(25\%)}  \\
head              & 0.643 (100\%) & 0.211 (25\%)*  & 0.537 (100\%)*     & 0.647 (100\%)   & 0.646 (25\%)   & 0.670 (25\%)   & \textbf{0.690} \red{(10\%)}  \\
shin              & 0.781 (100\%) & 0.375 (100\%)* & 0.628 (100\%)*     & 0.784 \red{(10\%)}    & 0.765 (75\%)*  & \textbf{0.81} \red{(10\%)}   & \textbf{0.81} \red{(10\%)} 
\\chest             & 0.715 (100\%) & 0.228 (50\%)*  & 0.650 (100\%)*     & 0.478 (75\%)*   & 0.720 (25\%)   & \textbf{0.722} \red{(10\%)}   & 0.716 (25\%)  \\
thigh             & 0.701 (100\%) & 0.283 (100\%)* & 0.586 (100\%)*     & 0.695 (75\%)*   & 0.656 (25\%)*  & 0.675 (75\%)*  & 0.690 (25\%)* \\
upper arm          & 0.726 (100\%) & 0.268 (75\%)*  & 0.595 (100\%)*     & 0.739 (75\%)    & 0.708 (100\%)* & \textbf{0.753} \red{(10\%)}   & 0.740 (25\%)  \\
waist             & 0.745 (100\%) & 0.297 (25\%)*  & 0.674 (100\%)*     & 0.582 (75\%)*   & 0.770 \red{(10\%)}   & 0.778 \red{(10\%)}   & \textbf{0.781} \red{(10\%)}  \\
        \hline
		\multicolumn{8}{l}{\textbf{Opportunity}} \\
		\hline
back              & 0.439 (100\%) & 0.164 (50\%)*  & 0.253 (25\%)*      & 0.446 \red{(10\%)}    & 0.445 (50\%)   & 0.380 (25\%)*  & \textbf{0.556} \red{(10\%)}  \\
lla               & 0.370 (100\%) & 0.197 (100\%)* & 0.398 (25\%)       & 0.386 (25\%)    & 0.375 (25\%)   & 0.374 (100\%)  & \textbf{0.516} \red{(10\%)}  \\
left shoe             & 0.391 (100\%) & 0.164 (10\%)*  & 0.396 (75\%)       & 0.282 (100\%)*  & 0.172 (25\%)*  & 0.164 (100\%)* & \textbf{0.416} \red{(25\%)}  \\
right shoe             & 0.378 (100\%) & 0.164 (10\%)*  & 0.392 (25\%)       & 0.265 (100\%)*  & 0.166 (50\%)*  & 0.183 (100\%)* & \textbf{0.402} \red{(10\%)}  \\
rua               & 0.416 (100\%) & 0.164 (10\%)*  & 0.293 (100\%)*     & 0.447 \red{(10\%)}    & 0.375 (10\%)*  & 0.277 (10\%)*  & \textbf{0.538} \red{(10\%)}  \\
        \hline
		\multicolumn{8}{l}{\textbf{PAMAP2 - Locomotion}} \\
		\hline
ankle             & 0.731 (100\%) & 0.609 (50\%)*  & 0.589 (10\%)*      & 0.651 (10\%)*   & 0.770 \red{(10\%)}   & 0.774 (50\%)   & \textbf{0.784} (100\%) \\
chest             & 0.654 (100\%) & 0.295 (50\%)*  & 0.738 \red{(10\%)}       & 0.669 (75\%)    & 0.655 \red{(10\%)}   & 0.730 \red{(10\%)}   & \textbf{0.741} \red{(10\%)}  \\
hand              & 0.723 (100\%) & 0.496 (25\%)*  & 0.731 (25\%)       & 0.723 (100\%)   & 0.750 \red{(10\%)}   & \textbf{0.791} \red{(10\%)}   & 0.740 \red{(10\%)}  \\
		\hline
		\multicolumn{8}{l}{\textbf{PAMAP2 - ADL}}\\
		\hline
		ankle & 0.550 (100\%) & 0.262 (100\%)* & 0.548 (100\%)* & {0.56} \red{(25\%)} & 0.489 (100\%)* & 0.567 (50\%) & \textbf{0.578} \red{(25\%)} \\
chest & 0.640 (100\%) & 0.156 (100\%)* & 0.64 (50\%) & {0.623} \red{(25\%)} & 0.607 (75\%)* & 0.615 (100\%)* & \textbf{0.651} (50\%) \\
hand  & 0.575 (100\%) & 0.208 (50\%)* & 0.585 \red{(25\%)} & 0.577 (75\%) & 0.586 (50\%) & 0.596 (50\%) & \textbf{0.617} \red{(25\%)} \\
    \hline
	\end{tabularx}
    \caption{\revisions{Classification performance: average of macro $F_1$ scores and the minimum percentage of labeled data outperforming the fully supervised model (Supervised-single). For each anchor device, bold numbers represent the highest $F_1$ score, and red numbers indicate the technique which requires the least amount of labeled data to outperform Supervised-single. When a technique does not outperform Supervised-single, we denote its best-achieved performance with an asterisk.}}
	\label{table:performance_device}
    \vspace{-0.4cm}
\end{table}

To study the low-data regime, we fine-tune \solution{} and other baselines except Supervised-single, using 10\%, 25\%, 50\%, 75\%, and 100\% of the available labeled data from the anchor device. The fine-tuned models are then evaluated on the anchor device from the validation/held-out group. Supervised-single is used as a reference point, thus trained using 100\% of the training data. Note that the labeled training data available in our datasets for each anchor device (on average) is as follows: \textsc{RealWorld}: 1027 windowed samples (approximately 51 minutes), \textsc{Opportunity}: 3014 samples (approximately 100 minutes), {\textsc{PAMAP2 - Locomotion}}: 1280 samples (approximately 42 minutes){, and \textsc{PAMAP2 - ADL}: 5709 samples (approximately 190 minutes)}. 

Table~\ref{table:performance_device} shows the classification performance for the various anchor devices, averaged over all validation groups in a leave-one-group-out evaluation. More specifically, we report the minimum percentage of labeled data required by each technique to surpass the performance of Supervised-single (in parenthesis), and the corresponding macro-$F_1$ score averaged over all validation groups. In case a technique does not surpass the performance of Supervised-single, we report its best performing $F_1$ score and labeled data percentage.  

Our results confirm the data-efficiency of \solution{}. \revisions{In 15 out of the 18 anchor devices, including those for ADL recognition}, \solution{} with 10\% or 25\% of labeled data achieves higher $F_1$ score than Supervised-single trained with 100\% labeled data. \revisions{In the remaining three cases}, \solution{} shows comparable $F_1$ score with 25\% of labeled data when evaluated at the thigh-worn device in RealWorld (\solution: 0.690,  Supervised-single: 0.701), a higher $F_1$ score with 100\% of labeled data when evaluated at the ankle-worn device in PAMAP2 (\solution: 0.784, Supervised-single: 0.731), \revisions{and a comparable $F_1$ score with 50\% of labeled data evaluated at the chest-worn device in PAMAP2 - ADL (\solution{}: 0.651, Supervised-single: 0.640)}.

Table~\ref{table:performance_device} also shows (in {red} font) the technique which requires the least amount of labeled data to surpass Supervised-single. We observe that \solution{} generally performs better compared with other semi-supervised approaches (AutoEncoder-single and AutoEncoder-multi) and self-supervised approach (Multi-task SSL). More specifically, \revisions{in 13 out of 18} anchor devices, \solution{} used the lowest percentage of labeled data across the baselines. This is remarkable considering that AutoEncoder-single, AutoEncoder-multi, and Multi-task SSL win at \revisions{5, 4, and 6} anchor devices; note that multiple winners can be chosen.   

Finally, Table~\ref{table:performance_device} shows (in bold font) the technique which provides the highest performance in a low-data regime. Here \solution{} has the highest $F_1$ score in \revisions{14 out of 18} anchor devices across all techniques. \revisions{Multi-task SSL also outperformed the supervised baseline in most cases, and outperformed \solution{} in 3 of the remaining scenarios. This could be attributed to the data-efficiency of self-supervised methods, and in certain scenarios, pre-text tasks based on specific data transformations could offer the right data diversity for training a feature extractor. However, for \emph{Opportunity} dataset, there are several cases where the Multi-task SSL baseline failed to outperform the supervised baseline. One possible reason for this is that the effectiveness of SSL methods which rely on manual data transformations can be highly dependent on the dataset and the specific transformations used by the technique~\cite{simclr_chen2020simple}. It is possible that the manual transformations employed by this baseline were not optimal for \textsc{Opportunity}, which resulted in poor recognition accuracy. Instead, \solution{} does not define any manual transformations on the datasets and leverages natural transformations present in the data; hence overall it is more robust to dataset variations.} \\

\begin{figure}[t]
    \begin{center}
        \begin{tabularx}{0.95\textwidth}{cc}
            \includegraphics[width=0.4\textwidth]{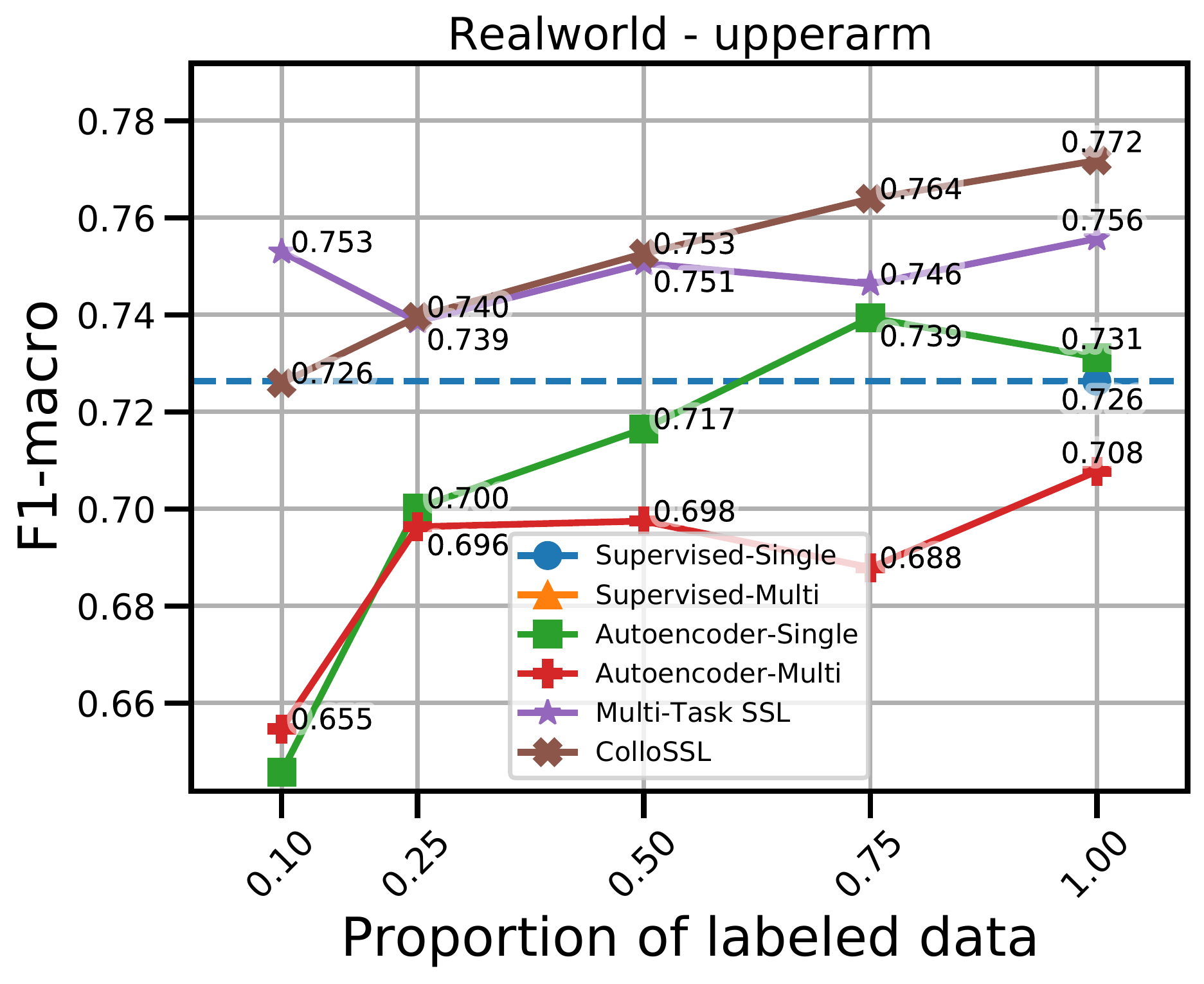}
            & \includegraphics[width=0.4\textwidth]{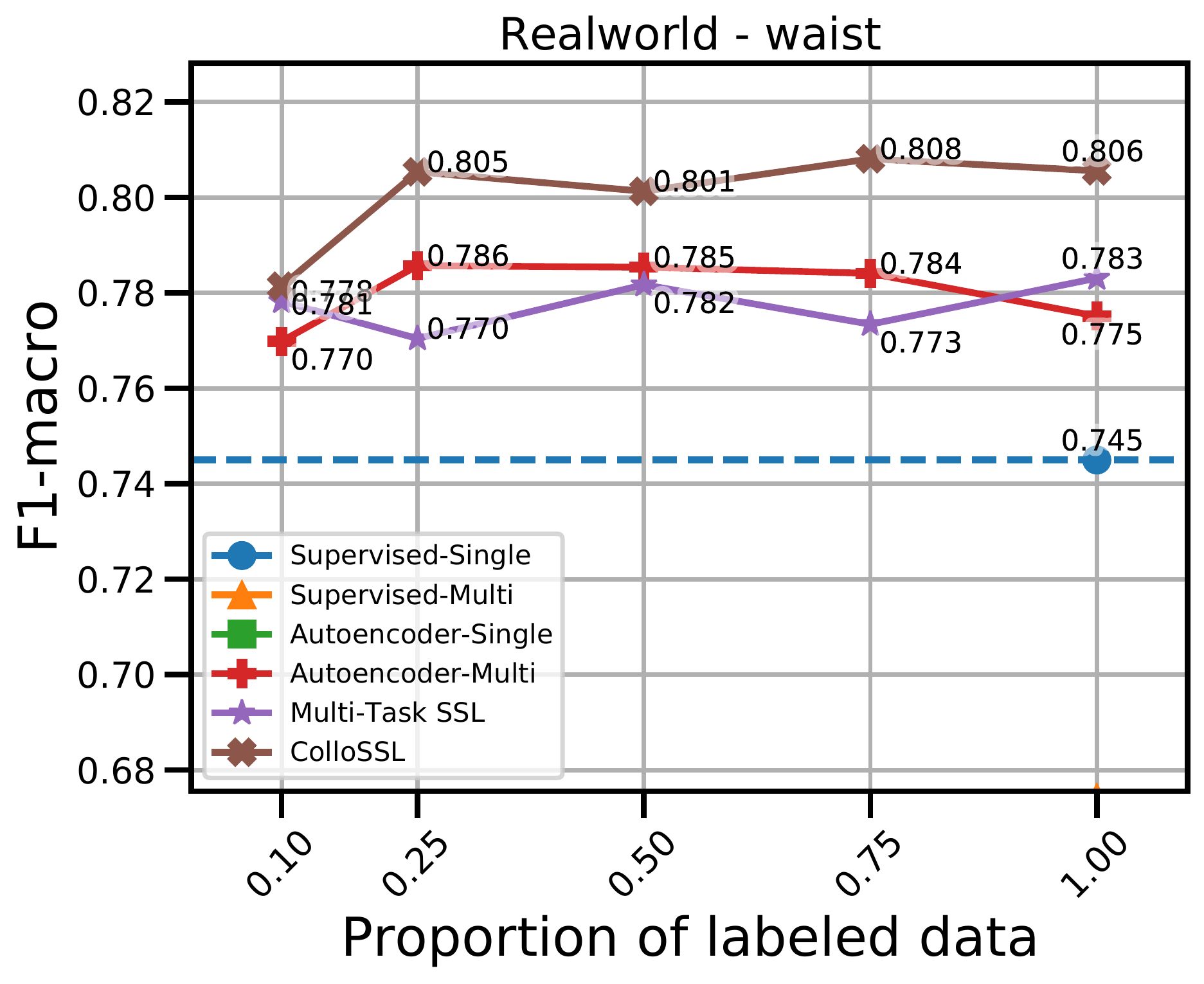} \\
            \includegraphics[width=0.4\textwidth]{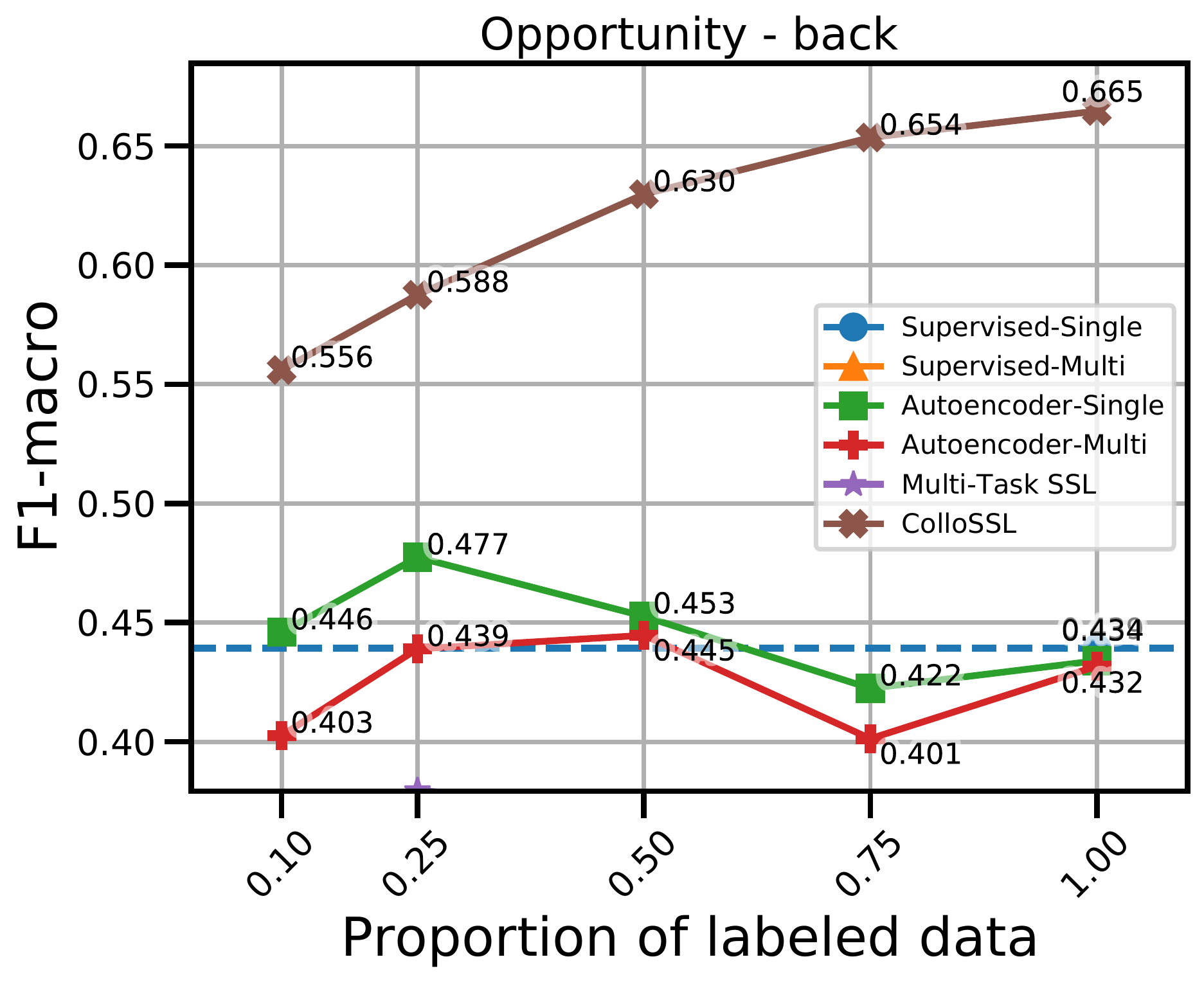}
            & \includegraphics[width=0.4\textwidth]{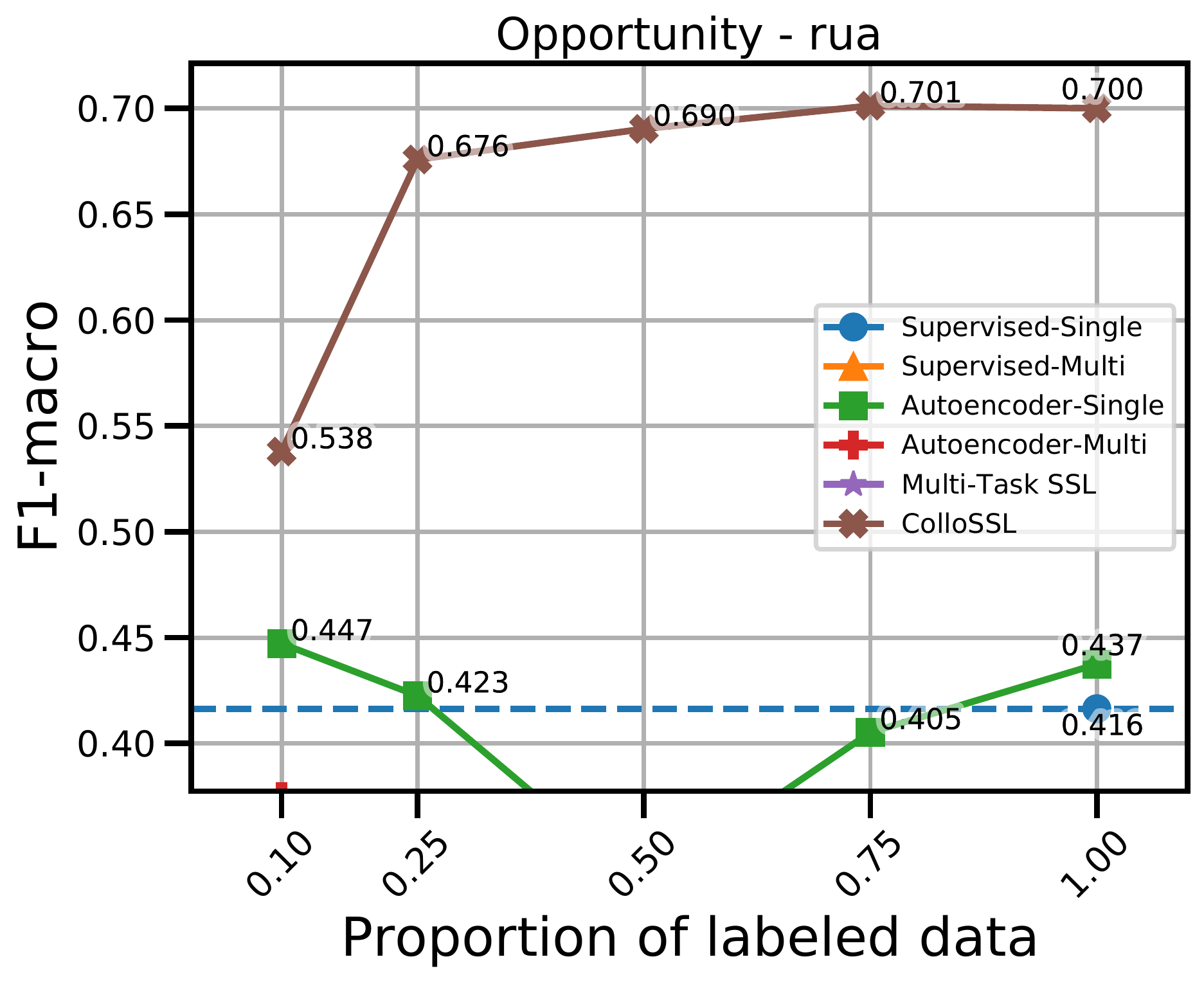} \\
        \end{tabularx}
    \caption{Assessing the classification performance of \solution{} and baselines across different percentages of labeled data. Note that some baselines had very poor performance and are not visible in the plot. Please refer to Table~\ref{table:performance_device} for in-depth results.}
    \label{fig:plot_proportion}
    \end{center}
    \vspace{-0.2cm}
\end{figure}

Figure~\ref{fig:plot_proportion} provides further insights into these findings by plotting the performance of various techniques when they trained or fine-tuned with different percentage of labeled data from the anchor device. We present two important findings. First, regardless of the percentage of labeled data used for fine-tuning, \solution{} generally outperforms the baselines. This shows that our design of device selection, contrastive sampling, and group contrastive loss contributes to enhancing the performance of HAR. Second, we can again observe that \solution{} outperforms the fully-supervised model (Supervised-single) even with much less labeled data \revisions{, including for PAMAP2 - ADL, which contains with more complex activities of daily living}.

\begin{table}[t]
	\small
	\begin{tabularx}{\textwidth}{p{0.1\textwidth}|p{0.1\textwidth}p{0.1\textwidth}p{0.1\textwidth}p{0.1\textwidth}p{0.1\textwidth}p{0.11\textwidth}p{0.11\textwidth}}	
		\hline
		\specialcell{Dataset\\(anchor)} & Random & \specialcell{Supervised-\\single} & \specialcell{Supervised-\\multi} & \specialcell{AutoEncoder-\\single} & \specialcell{AutoEncoder-\\multi} & \specialcell{Multi-task\\SSL} & \solution{} \\
		\hline \hline
		\multicolumn{8}{l}{\textbf{RealWorld}} \\
		\hline
		forearm   & 0.248 (0.028) & 0.732 (0.065)     & 0.495 (0.039)  & 0.723 (0.045)      & 0.718 (0.064)   & 0.738 (0.057)  & \textbf{0.774 (0.053)} \\
        head      & 0.123 (0.028) & 0.643 (0.055)     & 0.537 (0.031)  & 0.647 (0.071)      & 0.627 (0.061)   & 0.663 (0.026)  & \textbf{0.730 (0.046)} \\
        shin      & 0.375 (0.045) & 0.781 (0.044)     & 0.628 (0.060)  & 0.799 (0.064)      & 0.761 (0.078)   & 0.785 (0.052)  & \textbf{0.806 (0.103)} \\
        chest     & 0.135 (0.030) & 0.715 (0.104)     & 0.650 (0.054)  & 0.461 (0.127)      & \textbf{0.729 (0.09)}   & 0.708 (0.061)  & 0.720 (0.095) \\
        thigh     & 0.283 (0.024) & \textbf{0.701 (0.11)}     & 0.586 (0.022)  & 0.670 (0.061)      & 0.616 (0.088)   & 0.651 (0.120)  & {0.679 (0.101)} \\
        upper arm  & 0.126 (0.028) & 0.726 (0.090)     & 0.595 (0.019)  & 0.731 (0.066)      & 0.708 (0.063)   & 0.756 (0.084)  & \textbf{0.772 (0.042)} \\
        waist     & 0.157 (0.049) & 0.745 (0.127)     & 0.674 (0.042)  & 0.579 (0.167)      & 0.775 (0.051)   & 0.783 (0.102)  & \textbf{0.806 (0.070)} \\
        \hline
        Average & 0.207 (0.090) & 0.720 (0.039) & 0.595 (0.058) & 0.659 (0.103) & 0.705 (0.057) & 0.726 (0.050) & \textbf{0.755 (0.044)}\\

        \hline
		\multicolumn{8}{l}{\textbf{Opportunity}} \\
		\hline
        back  & 0.164 (0.010) & 0.439 (0.092)     & 0.217 (0.012)  & 0.434 (0.114)      & 0.432 (0.107)   & 0.355 (0.076)  & \textbf{0.665 (0.134)} \\
        lla   & 0.197 (0.050) & 0.370 (0.013)     & 0.396 (0.082)  & 0.458 (0.028)      & 0.369 (0.016)   & 0.374 (0.011)  & \textbf{0.553 (0.018)} \\
        left shoe & 0.164 (0.009) & 0.391 (0.043)     & 0.394 (0.046)  & 0.282 (0.073)      & 0.171 (0.010)   & 0.164 (0.009)  & \textbf{0.443 (0.040)} \\
        right shoe & 0.164 (0.009) & 0.378 (0.024)     & 0.354 (0.032)  & 0.265 (0.056)      & 0.164 (0.009)   & 0.183 (0.011)  & \textbf{0.448 (0.026)} \\
        rua   & 0.164 (0.009) & 0.416 (0.060)     & 0.293 (0.068)  & 0.437 (0.126)      & 0.277 (0.058)   & 0.185 (0.034)  & \textbf{0.700 (0.131)} \\
        \hline
        Average & 0.171 (0.013) & 0.399 (0.025) & 0.331 (0.068) & 0.375 (0.084) & 0.283 (0.106) & 0.252 (0.092) & \textbf{0.562 (0.107)} \\

        \hline
		\multicolumn{8}{l}{\textbf{PAMAP2 - Locomotion}} \\
		\hline
        ankle & 0.558 (0.115)  & 0.731 (0.100)     & 0.558 (0.072)  & 0.635 (0.012)      & 0.754 (0.081)   & 0.720 (0.095)  & \textbf{0.784 (0.088)} \\
        chest & 0.160 (0.052) & 0.654 (0.136)     & 0.680 (0.082)  & 0.687 (0.120)      & 0.639 (0.098)   & 0.716 (0.141)  & \textbf{0.742 (0.112)} \\
        hand  & 0.397 (0.180) & 0.723 (0.111)     & 0.738 (0.092)  & 0.723 (0.105)      & 0.729 (0.088)   & \textbf{0.777 (0.065)}  & 0.737 (0.078) \\
		\hline
        Average      & 0.372 (0.163) & 0.703 (0.121)     & 0.659 (0.111)  & 0.682 (0.099)      & 0.708 (0.102)   & 0.738 (0.109)  & \textbf{0.754 (0.097)} \\
        \hline
		\multicolumn{8}{l}{\textbf{PAMAP2 - ADL}} \\
		\hline
        ankle & 0.262 (0.038) & 0.550 (0.124) & 0.548 (0.135) & 0.566 (0.090) & 0.489 (0.151) & 0.559 (0.096) & \textbf{0.646 (0.184)}\\
        chest & 0.156 (0.062) & 0.640 (0.189) & 0.655 (0.177) & \textbf{0.66 (0.150)} & 0.606 (0.104) & 0.615 (0.169) & \textbf{0.66 (0.185)} \\
        hand & 0.170 (0.030) & 0.575 (0.078) & 0.647 (0.090) & 0.575 (0.063) & 0.585 (0.095) & 0.621 (0.089) & \textbf{0.664 (0.087)} \\
        \hline
        Average & 0.196 (0.047) & 0.588 (0.038) & 0.617 (0.049) & 0.601 (0.044) & 0.560 (0.051) & 0.598 (0.028) & \textbf{0.657 (0.008)} \\
        \hline
        \hline
        \specialcell{\textbf{Total}\\\textbf{average}}    & 0.237 & 0.603 & 0.551 & 0.579 & 0.564 & 0.579 & \textbf{0.682}          \\
        \hline
	\end{tabularx}
	\caption{\revisions{Comparison of classification performance (average and standard deviation of macro $F_1$ scores) for different anchor devices, when 100\% of the labeled data is available for fine-tuning. lla: left lower arm, rua: right upper arm.}}
	\label{table:performance_device_all}
    \vspace{-0.6cm}
\end{table}

\subsection{Comparison of \solution{} Recognition Performance with Baselines}

We now compare the classification performance of \solution{} against various baseline techniques when sufficient labeled data is available. Here, we use 100\% of the labeled training data available from the anchor device for fine-tuning \solution{}, AutoEncoder-single, AutoEncoder-multi, and Multi-task SSL, and for training Supervised-single and Supervised-multi. Then, we evaluate these techniques on the anchor device from the held-out group. A hyperparameter search on training parameters was conducted for all pipelines to ensure optimal performance.

Table~\ref{table:performance_device_all} shows the mean and standard deviation of macro $F_1$ scores for different anchor devices. On average, the results show that \solution{} outperforms baseline techniques for all datasets we used. \solution{} has an absolute increase of around \revisions{7.9\% in the $F_1$ score over the average of all anchor devices across all datasets}, compared to the best performing baseline, Supervised-single. We also observe that \solution{} outperforms Multi-task SSL, a state-of-the-art self-supervised learning technique for HAR for all except one anchor device. This validates our design choice of leveraging natural transformations from the \problem{} settings for self-supervised contrastive learning. \revisions{Furthermore, our method outperforms the best performing baseline by 4\% in $F_1$ score in absolute terms in PAMAP2 - ADL, which demonstrates that our proposed method can offer performance gain in simpler locomotion recognition, as well as more complex ADL recognition.}

\subsection{Embedding Similarity and Data Saliency}

In this section, we delve deeper to analyze the feature embeddings learned by the feature extractor and compare them between \solution{} and Fully Supervised settings. We also present Saliency Maps to understand how the HAR models trained in these two settings are making their predictions. 

\begin{figure}[t]
    \begin{center}
        \begin{tabularx}{\textwidth}{cc}
            \includegraphics[width=\textwidth]{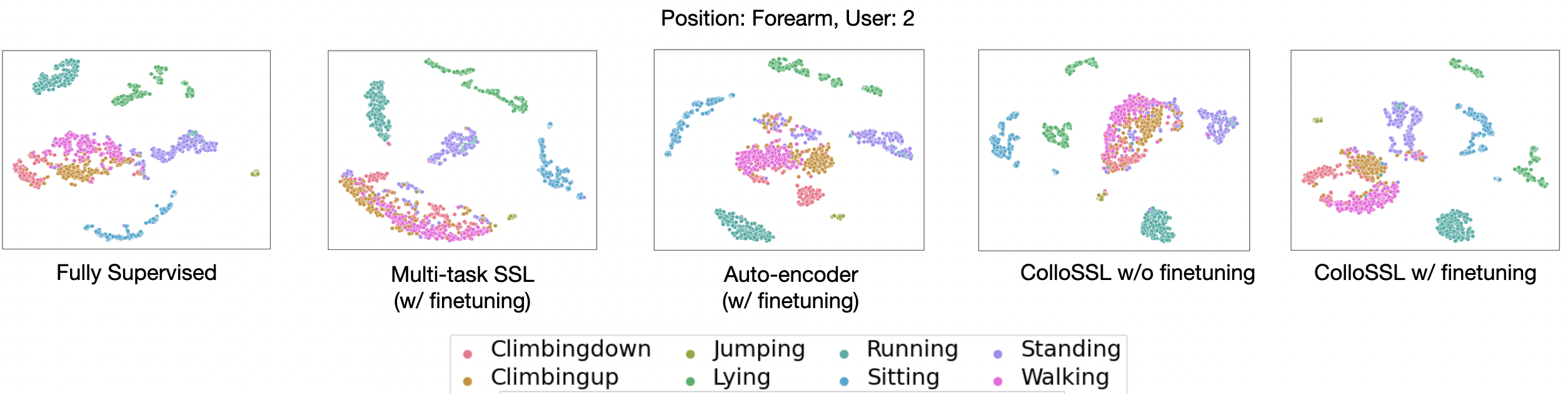}\\
            \includegraphics[width=\textwidth]{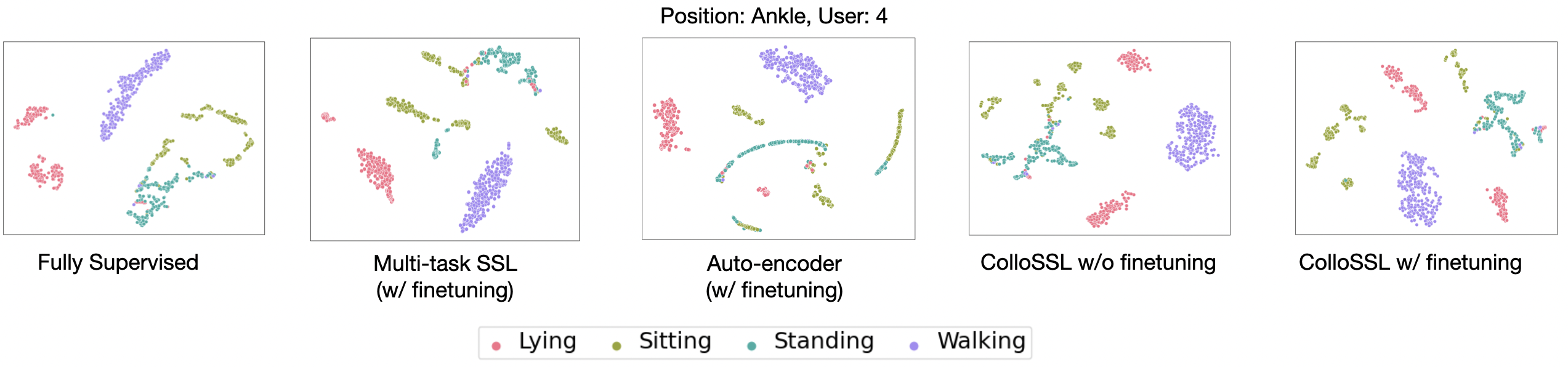} \\
        \end{tabularx}
\caption{t-SNE visualizations to compare the features learned by \solution{} and various baselines.}
    \vspace{0.1cm}
\label{fig:tsne}
    \end{center}
\end{figure}

\parjump{}
\noindent
\textbf{Visualizing the feature space using t-SNE plots:} We visualize the learned feature embeddings of \solution{} and the baselines (\textsc{Supervised-single}) using t-distributed stochastic neighbor embedding (t-SNE) plots~\cite{van2008visualizing}. t-SNE is a statistical method that is used to explore and visualize high-dimensional data. Based on Stochastic Neighbor Embedding, it reduces the dimensionality of data in a nonlinear way and places each data point in a location in a space of two or three dimensions. Specifically, t-SNE aims to discover patterns in the data by clustering data points based on their similarity. 

\revisions{Using t-SNE, we project the 96-dimensional feature embeddings generated by the feature extractor onto a 2D space in the following settings: \solution{} w/o finetuning, \solution{} w/ finetuning, and fully-supervised, multi-task SSL and autoencoder-single.} \revisions{Figure~\ref{fig:tsne} shows the t-SNE plots for two users and two anchor devices from the RealWorld (top) and PAMAP2 (bottom) datasets. In common, we observe that \solution{} w/o finetuning already generates well-separable feature embeddings across classes. It implies that \solution{} captures the semantic structure of the data very well. The class separation in the embeddings is further improved by finetuning with a small amount of labeled data as shown in \solution{} w/ finetuning. We also observe that the nature of clustering learned with \solution{} is largely comparable with those learned with fully-supervised model trained with 100\% of the labeled data. The two baselines, autoencoder and multi-task SSL, also achieve a reasonably good cluster separation; however certain classes end up having high overlap in their features. For example, multi-task SSL finds it difficult to separate the data \emph{Climbing up, Climbing Down, and Walking} activities in Figure \ref{fig:tsne} (top).}

\begin{figure}
  \subfloat[Climbing down activity collected from a chest-work IMU]{
	\begin{minipage}[c][1\width]{
	   0.48\textwidth}
	   \centering
	   \includegraphics[width=\linewidth]{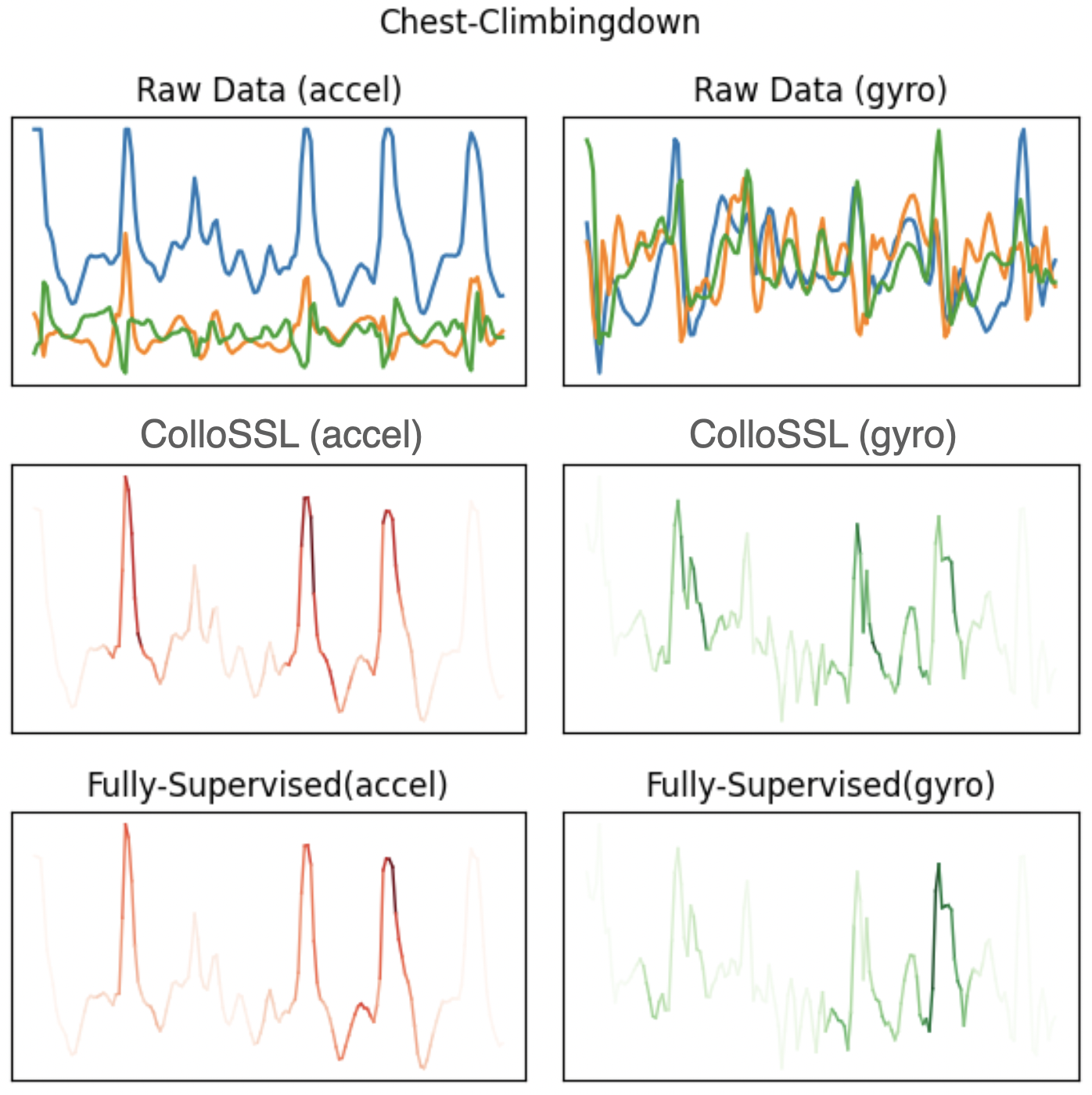}
	\end{minipage}
	\label{fig:saliency_map_chest}}
    \hfill
  \subfloat[Walking activity collected from a waist-worn IMU]{
	\begin{minipage}[c][1\width]{
	   0.48\textwidth}
	   \centering
	   \includegraphics[width=\linewidth]{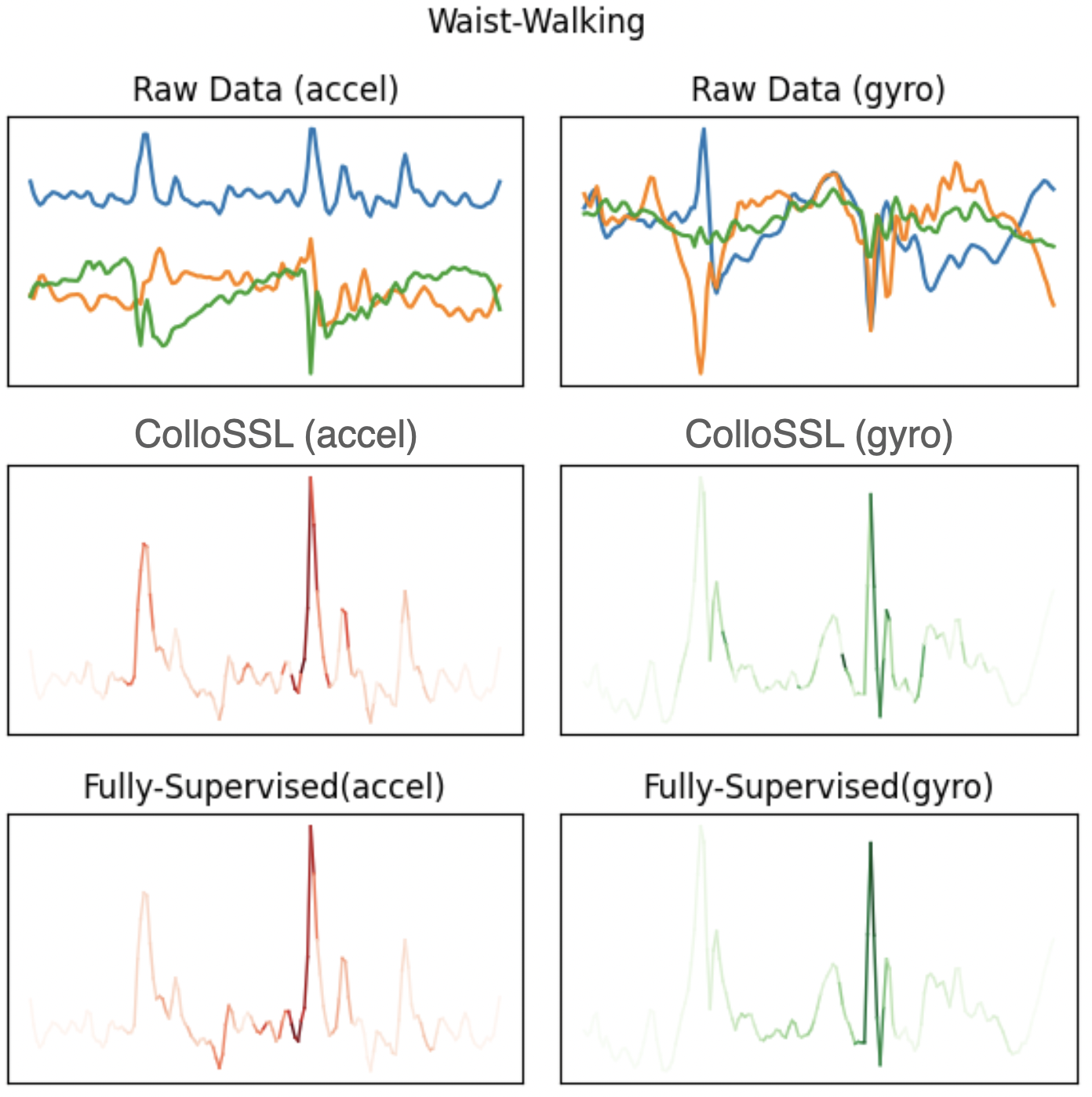}
	\end{minipage}
	\label{fig:saliency_map_waist}}
\caption{Saliency map for samples of RealWorld dataset; (top) raw input signal, (middle) and (bottom) magnitude values computed from the input signal. The intensity of color indicates the impact of the region on the model prediction. We observe that \solution{} and fully supervised model show similar patterns of the color intensity. This implies that models trained with both approaches largely focus on similar regions of the data to make their predictions.}
\label{fig:saliency_map}
\vspace{-0.3cm}
\end{figure}

\parjump{}
\noindent
\textbf{Visualizing the salient regions in the data using Saliency maps}. For a better interpretability of our findings, we visualize saliency maps~\cite{simonyan2013deep, saeed2019multi} for two randomly selected data samples from the \textsc{RealWorld} dataset. A saliency map illustrates the regions of the data sample that have the most effect on a model's prediction; the parts with higher color intensity shows the regions that contribute most to the model’s prediction. Our objective is to understand if the salient regions of the data remain consistent across \solution{} and Fully-Supervised training. 

Figure~\ref{fig:saliency_map} shows the saliency maps for two randomly selected data samples from the RealWorld dataset; a sample of a \textit{climbing down} activity collected from a chest-worn IMU in Figure~\ref{fig:saliency_map_chest} and a sample of a \textit{walking} activity collected from a waist-worn IMU. We visualize the three-axis raw input data from the accelerometer and gyroscope separately in the top pane. The middle and bottom panes show the saliency maps for this input data produced by \solution{} and Fully-supervised training for a class with the highest prediction score. For ease of understanding, we only present the magnitude values of the accelerometer and gyroscope data in the saliency maps. In the middle and bottom panes, the intensity of color indicates the impact of the region on the model prediction. The regions with strong intensity imply that they contribute to the model prediction more than those with weak intensity. 

Figure~\ref{fig:saliency_map} shows that \solution{} and the fully-supervised model show a similar pattern in color intensity, both for accelerometer and gyroscope samples. For example, in Figure~\ref{fig:saliency_map_chest}, the periodic peaks in the accelerometer data on the x-axis (blue color) are largely responsible for the model's prediction in both \solution{} and fully-supervised settings. The takeaway from this result is that the models trained with \solution{} and fully-supervised training largely focus on similar regions of the data to make their predictions. This confirms that \solution{} is able to generate meaningful representations of data for the HAR classification task. 

\section{Analysis}
\label{sec:data_analysis}
In this section, we present a set of ablation studies and also analyze the performance of \solution{} under real-world constraints in sensor devices. \revisions{Please note that due to space constraints, we present the results on only one dataset and a subset of anchor devices in each section, however the results also hold for other scenarios.}   

\subsection{\revisions{Analysis of the Device Selection Strategy}}
\label{subsec:selection_strategies}
\revisions{Our proposed device selection strategy (\S\ref{subsec:device_selection}) uses the closest device (least MMD distance) to the anchor as the Positive device and all devices as Negatives, weighted by the inverse of their MMD distance to anchor. In this ablation experiment, we compare this strategy against two baselines: i) Random Selection ii) Closest Positive and Random Negative Selection. In the Random Selection strategy, we randomly pick one positive and one negative with replacement in each batch. In the Closest Positive \& Random Negative strategy, we pick the closest device with the least MMD distance to anchor, but randomly choose one negative device.} 

\revisions{Table \ref{table:random_selection} shows the experiment result on the \textsc{PAMAP2 - ADL} dataset with 12 ADL activities. We observe that \solution{} outperforms the two baseline approaches. In particular, the Random Selection strategy performs the worse as it often picks positive devices which have different data distributions from the anchor device. The Closest Positive \& Random Negative also has a lower performance as it does not prevent the violation of the \textbf{(N1)} criteria for negative samples as described in \S\ref{sec.dds-cs}. This finding supports our hypothesis that using `all' devices for negative samples serve as a hedge against the scenario when \textbf{(N1)} is violated on one of the devices.}

\begin{table}[t]
	\small
	\begin{tabular}{p{0.1\textwidth}|ccc}
		\hline
Anchor        & Closest Positve \& Random Negative     & Random Selection  & \solution{}           \\
		\hline \hline
chest           & 0.649 (0.175) & 0.631 (0.166) & \textbf{0.662 (0.185)} \\
ankle           & 0.602 (0.122) & 0.553 (0.095) & \textbf{0.646 (0.184)} \\
hand            & 0.651 (0.088) & 0.634 (0.085) & \textbf{0.664 (0.087)} \\
		\hline
	\end{tabular}
	\caption {\revisions{Comparison of various device selection strategies in terms of performance (average and standard deviation of macro $F_1$ scores) for the \textsc{PAMAP2 - ADL} dataset.}}
	\label{table:random_selection}
    \vspace{-0.4cm}
\end{table}

\subsection{\revisions{Analysis of the Contrastive Sampling Approach}}
\label{subsec:contrastive_sampling_experiment}
\revisions{We now evaluate the effect of contrastive sampling by running an ablation on the \textsc{PAMAP2-ADL} dataset. We compare \solution{}'s asynchronous negative sampling against its counterpart, synchronous negative sampling. Note that, positive samples are sampled synchronously in both the settings, otherwise positive samples will violate characteristic \textbf{(P1)}. Table~\ref{table:contrastive_sampling} exhibits the improvement in performance by using asynchronous negative sampling. We attribute this gain to the knowledge that synchronous samples in \problem{} setting belongs to the same class as the anchor sample; hence by negatively contrasting these samples,  the feature extractor is violating \textbf{(N1)} and learns poor representations.}
\begin{table}[t]
	\small
	\begin{tabular}{p{0.1\textwidth}|cc}
		\hline
 Anchor         & Synchronous positive and negative    & \solution{}         \\
		\hline \hline
chest           &   0.6307 (0.180) & \textbf{0.662 (0.185)} \\
ankle           &    0.601 (0.101) & \textbf{0.646 (0.184)}	\\
hand            &   0.639 (0.076) &   \textbf{0.664 (0.087)} \\
		\hline
	\end{tabular}
	\caption {\revisions{Comparison of the effect of contrastive sampling on performance (average and standard deviation of macro $F_1$ scores) for the \textsc{PAMAP2 - ADL} dataset.}}
	\label{table:contrastive_sampling}
    \vspace{-0.4cm}
\end{table}

\begin{table}[t]
	\small
	\begin{tabular}{p{0.1\textwidth}|cc}
		\hline
 Anchor         & \solution{} w/o weights    & \solution{}           \\
		\hline \hline
chest           &   0.658 (0.184) & \textbf{0.662 (0.185)} \\
ankle           &    0.601 (0.112) & \textbf{0.646 (0.184)}	\\
hand            &   0.636 (0.076) &   \textbf{0.664 (0.087)} \\
		\hline
	\end{tabular}
	\caption {\revisions{Comparison of the effect of Weights over performance (average and standard deviation of macro $F_1$ scores) of \solution{} in the \textsc{PAMAP2 - ADL} dataset.}}
	\label{table:weightsVsNon-weights}
    \vspace{-0.4cm}
\end{table}

\subsection{\revisions{Analyzing the Role of Weights in Multi-view Contrastive Loss}}
\label{subsec:weightsVsNon-weights}
\revisions{In multi-view contrastive loss, we introduce weights as a way to differentiate between the contributions of each negative device towards the optimization objective. To investigate the effect these weights have on \solution{}, we conducted an experiment where the weights were removed from the loss function (i.e., all devices were assigned the same weight). Table \ref{table:weightsVsNon-weights} shows the result of our experiment. We can observe that use of the weighted loss criterion improves the performance of \solution{} which supports our hypothesis that weighted negatives help in pushing closer negative embeddings further away from the anchor device \textbf{(N2)} resulting in better features.}

\subsection{\revisions{Robustness to Sensor Heterogeneity}} 
\label{subsec:sensor_noise}
\revisions{In the \problem{} setting, devices placed at different body positions could be heterogeneous, in that they can come from different manufacturers or use different inertial sensors.} \revisions{In this section, we probe the robustness of \solution{} to sensor heterogeneity by synthetically adding two types of heterogeneity in the IMU data based on prior literature~\cite{frosio2008autocalibration, poddar2017comprehensive, Grammenos:2018:YSB:3200905.3191743}. }

\revisions{Prior research has shown that deterministic errors in IMU sensors are the prominent causes of heterogeneity in the sensor data. Deterministic errors are caused by variations in sensor components across manufacturers, imperfections introduced in the analog circuitry of the sensor during the manufacturing process~\cite{dey2014accelprint}, or temperature differences between initial calibration and operational stages~\cite{aggarwal2008standard}. Two of the major types of deterministic errors are scale factor errors and bias errors~\cite{Grammenos:2018:YSB:3200905.3191743}. }

\revisions{For this experiment, we induce different scale factor and bias errors to each IMU device in the \textsc{RealWorld} dataset. For each device, we sample a scale factor $S$ from a normal distribution with $\mu = 1.0$ and $\sigma = 0.05$ (low heterogeneity) and $\sigma = 0.1$ (high heterogeneity). Similarly, we sample a bias factor $B$ from a normal distribution with $\mu = 0.0$ and $\sigma = 0.05$ (low heterogeneity) and $\sigma = 0.1$ (high heterogeneity). Following the methodology proposed in \cite{Grammenos:2018:YSB:3200905.3191743}, the two factors are introduced to the raw datasets $\{\mathbb{X}_i\}_{i=1}^{D}$ to obtain $\mathbb{X'}_i = S \times \left(\mathbb{X}_i - B\right) $, where $\mathbb{X'}_i$ denotes the dataset with induced sensor heterogeneity for the $i^{th}$ device. Thereafter, we run the end-to-end training pipeline of \solution{} on the heterogeneous datasets $\{\mathbb{X'}_i\}_{i=1}^{D}$ using the same experiment protocol as previous experiments.}

\revisions{Our findings are shown in Table~\ref{table:sensor_noise}. For comparison, we also present the results when no additional sensor error is added to the  dataset. Overall, we observe that \solution{} is able to handle sensor heterogeneity and outperforms the fully-supervised and multi-task SSL baselines. Interestingly, we found that introducing `low heterogeneity' to the unlabeled data improves the performance of \solution{} and the other baselines. This finding can be explained by prior work which has shown that data augmentation based training of deep neural networks helps in learning better more generalizable features~\cite{park2019specaugment, mathur2018using}}. 

\begin{table}[t]
\small
\begin{tabular}{@{}c|ccc|ccc|ccc@{}}
\toprule
        & \multicolumn{3}{c|}{None}                                                                                                        & \multicolumn{3}{c|}{Low}                                                                                                        & \multicolumn{3}{c}{High}                                                                                                        \\ \midrule \midrule
Anchor  & \begin{tabular}[c]{@{}c@{}}Supervised\\single\end{tabular} & \begin{tabular}[c]{@{}c@{}}Multi-task \\ SSL\end{tabular} & \solution{} & \begin{tabular}[c]{@{}c@{}}Supervised\\single\end{tabular} & \begin{tabular}[c]{@{}c@{}}Multi-task \\ SSL\end{tabular} & \solution{} & \begin{tabular}[c]{@{}c@{}}Supervised\\single\end{tabular} & \begin{tabular}[c]{@{}c@{}}Multi-task \\ SSL\end{tabular} & \solution{} \\
\midrule
forearm & 0.732 & 0.738 & \textbf{0.774} & 0.744 & 0.762 & \textbf{0.786} & 0.73 & 0.74 & \textbf{0.761}   \\
shin    & 0.781 & 0.785 & \textbf{0.806} & 0.806 & 0.811 & \textbf{0.8337} & 0.773 & 0.782 & \textbf{0.80}   \\ \bottomrule
\end{tabular}
\caption{\revisions{Performance (macro $F_1$ scores) for two anchor devices in the \textsc{RealWorld} dataset under different levels of sensor heterogeneity. `None' denotes the case where no additional sensor error is added to the dataset.}} 
\label{table:sensor_noise}
    \vspace{-0.4cm}
\end{table}

\subsection{\revisions{Robustness to Missing Devices}}
\label{subsec:missing_data}

\revisions{In real-world scenarios, it is often the case that all the devices are not available all the time, e.g., the device runs out of battery, or a user takes off their earbuds during a conversation. This would result in having missing signal data from some devices in the \problem{} setting. We explore this missing data problem for \solution{} and conduct an experiment with \textsc{RealWorld} dataset. While preparing the data for this experiment, we assume that the anchor device, for which we would like to learn a prediction model, is always available. For the remaining $N$ devices, we set the unavailability of each device with the probability, $p_u$= \{0.1, 0.2, 0.3, 0.4, 0.5\}. For example, if $p_u$ is set to 0.1, all devices will be available in the same time window with the probability of $(1-0.1)^N$. More specifically, when a device is set to unavailable in a given time window, we replace its sensor data with zeros.}

\begin{figure}[t]
    \begin{center}
    \includegraphics[width=0.4\textwidth]{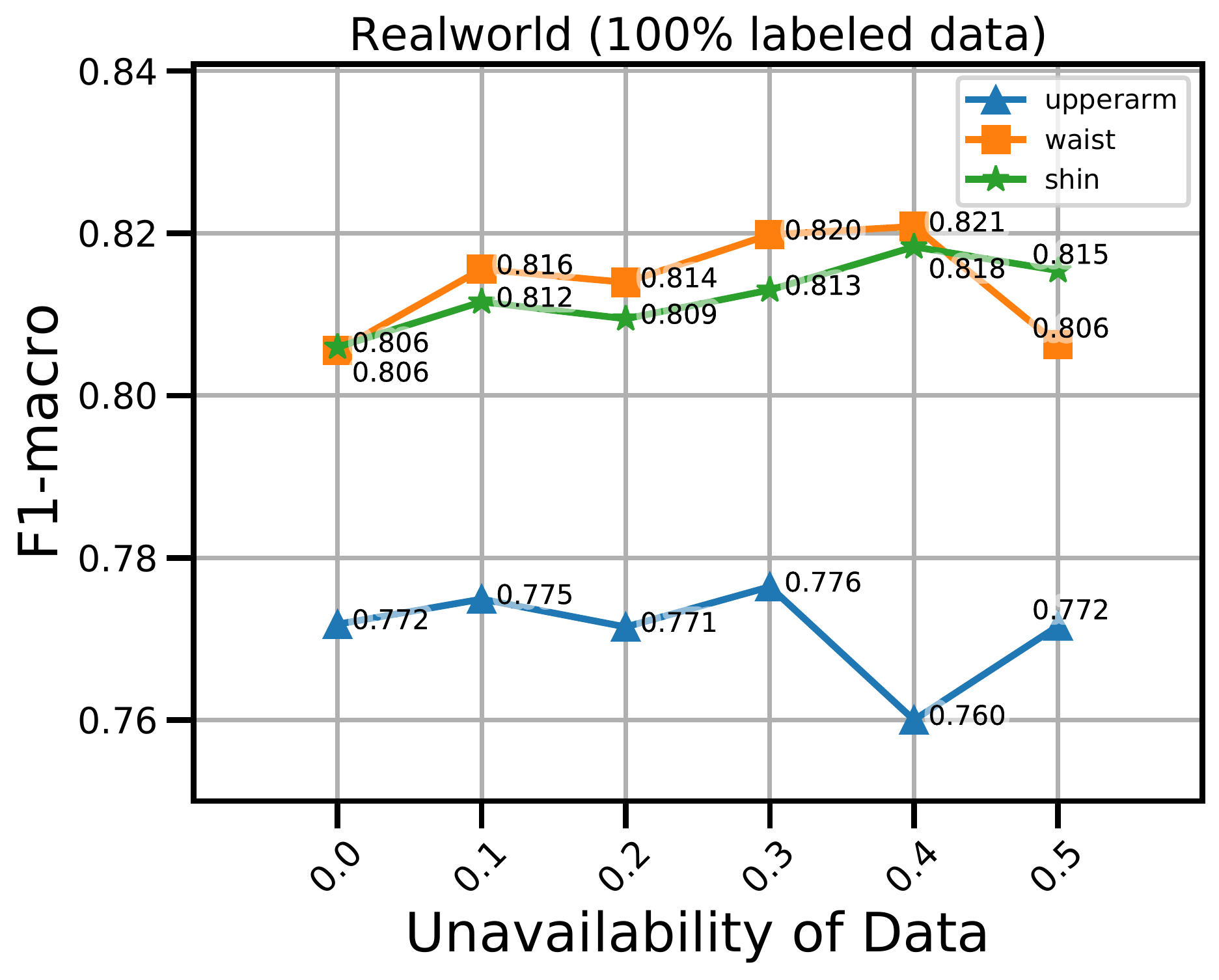}
        \vspace{-0.3cm}

    \caption{\revisions{Assessing the classification performance of \solution{} across different unavailability of devices in \problem{} setting. Note that unavailability of each device (x-axis) is decided independently of each other. Please refer to \S~\ref{subsec:missing_data} for more details.}}
    \label{fig:performance_missing_data}
    \end{center}
    \vspace{-0.1cm}
\end{figure}

\revisions{Fig~\ref{fig:performance_missing_data} shows the $F_1$-macro values with varying availability probability values. The results show that \solution{} is robust against the changing availability of devices and we observe at most a 1\% performance drop in our experiments due to device unavailability. This result can be explained by the design of device selection and weighted loss function in \solution{}. Firstly, missing devices (i.e., devices with 0 data) will have a high MMD with the anchor device, and \solution{}  is likely to assign them as negative devices. Secondly, the contribution of these missing devices will be significantly down-weighted in the multi-view loss function as negative devices with high MMD distances get assigned smaller weights in training. Surprisingly, we also observe that \solution{} with missing devices sometimes provides slightly higher performances than the case with full device availability. Admittedly, we do not have a good explanation for this behavior; we surmise that the neural network considers missing data as a form of noise, which might lead to an implicit training regularization that boosts performance.}

\subsection{Robustness to Temporal Misalignment}

\label{subsec:misalignment}
\begin{table}[t]
	\small
	\begin{tabular}{p{0.15\textwidth}|ccccccccc}
		\hline
		Anchor device & \multicolumn{8}{c}{Time Synchronization error}     \\
		\cline{2-9}
		(RealWorld)  & 0s  & 0.01s & 0.1s & 0.5s & 0.75s & 1.5s & 2.25s & 3s \\
		\hline \hline
waist & 0.806 & 0.804 & 0.800 & 0.808 & 0.811 & 0.805 & 0.802  & 0.812 \\
shin   & 0.806 & 0.809 & 0.809 & 0.805 & 0.807 & 0.813 & 0.811 & 0.815 \\
		\hline
	\end{tabular}
	\caption {\revisions{Comparison of classification performance (average of macro $F_1$ scores) for different time synchronization errors.}}
	\label{table:performance_noise}
    \vspace{-0.4cm}
\end{table}

\revisions{In \S\ref{subsec:tsmds}, we assume that data from multiple devices in the \problem{} setting are collected in a time-aligned manner. To investigate how robust \solution{} is to temporal misalignment between devices, we conduct an experiment with the RealWorld dataset by deliberately injecting time synchronization errors. More specifically, we shift the timestamps of all devices in the RealWorld dataset by 0.01, 0.1, 0.5, 0.75, 1.5, 2.25, and 3 seconds, except the anchor device.}

\revisions{Table~\ref{table:performance_noise} shows the $F_1$-macro values for two anchor devices, \textit{waist} and \textit{shin}. The results show that, for realistic, moderate time-sync errors ($\leq$ 0.5 seconds), there is no significant change in the performance of \solution{}, i.e., within \textpm 0.006 of the $F_1$-macro value. Even with high misalignment casees ($>$ 0.5 seconds), the change in the $F_1$-macro score is about \textpm 0.01. We conjecture that this result is caused by a) the temporal locality of human behaviors, and b) the ability of the feature extractor $f(.)$ to ignore moderate synchronization errors.}

\subsection{Generalizability of the Feature Extractor}
\label{subsec:generalizability}
\begin{table}[t]
	\small
	\begin{tabular}{p{0.2\textwidth}|cc}
		\hline
Device for testing         & \solution    & \solution-unseen           \\
		\hline \hline
upper arm       & 0.772 (0.042) & 0.764 (0.063) \\
waist           & 0.806 (0.070) & 0.792 (0.072) \\
		\hline
	\end{tabular}
	\caption {Comparison of classification performance (average and standard deviation of macro $F_1$ scores) between \solution{} and \solution-unseen in the RealWorld dataset.}
	\label{table:performance_generalize}
    \vspace{-0.4cm}
\end{table}

We further investigate the generalizability of \solution: whether the feature extractor $f(.)$ trained using \solution{} is transferable to new devices, i.e., the ones that do not participate in pre-training of $f(.)$. To this end, we pre-train \emph{\solution-unseen} on unlabeled data from all devices except one `unseen' device. The pre-trained model is then fine-tuned and evaluated on the unseen device. For example, in the RealWorld dataset -- if 'head' is chosen as the unseen device, we pre-train the feature extractor on rest of the devices to obtain \solution-unseen. Then, we fine-tune \solution-unseen with labeled data of 'head', and report the classification performance using test data of 'head'. 

Table~\ref{table:performance_generalize} compares the classification performance between \solution{} and \solution-unseen when the model is evaluated at upper arm- and waist-worn devices in the RealWorld dataset. The results show that \solution-unseen shows comparable performance to \solution, even though the data of the unseen device is not used for pre-training the feature extractor. The decrease of $F_1$ score is less than 1\% in both cases. This gives us an early indication that the feature extractor, $f(.)$, trained using \solution{} is transferable across devices and can be useful for finetuning on unseen devices. More specifically, when a new, unseen device is added to the \problem{} setting, we can reuse the pre-trained $f(.)$ and just fine-tune it using a small amount of labeled data from the new device.

\section{Discussion and Limitations}
\label{sec:limitations}
In this section, we discuss the limitations of our approach, elaborate on some of the practical deployment concerns associated with our method, and highlight avenues for future research on this topic.

\parjump{}
\noindent
\textbf{Training Cost of \solution{}.} \revisions{Training a model using \solution{} naturally takes more time when compared to fully-supervised learning, because of the need for pre-trainig on unlabeled data.} However, since model training is currently done offline (e.g., on a server), it has no adverse implications for system resources on mobile or wearable devices. Further, \solution{} does not impose any additional costs for data collection; in the \problem{} setting, multiple devices (e.g., smartphone, smartwatch) are anyway collecting sensor data related to a user's activity, and \solution{} simply uses this unlabeled data to train a more accurate HAR model. 

\parjump{}
\noindent
\textbf{Runtime System Cost.} Although \solution{} uses data from multiple devices to train the HAR model, it is important to note that the trained model using \solution{} only operates on a single device at runtime, similarly to any conventional HAR model. Hence, we expect that the system costs of \solution{}, such as inference latency and energy consumption on mobile and wearable devices, are the same as an HAR model trained using supervised learning.

\parjump{}
\noindent
\textbf{\solution{} as a general framework for learning in \problem{} settings.} Although we focus on applying \solution{} to HAR with motion data, the \problem{} setting is common to other sensor modalities (e.g., audio, vision) as shown in Figure~\ref{fig:tsmds}. To apply \solution{} to other \problem{} settings, the technical solutions (device selection, contrastive sampling, and group contrastive loss) need to be redesigned to reflect the characteristics of sensory signals, user behavior, and environments. However, we believe the key idea of \solution{} is still valid, which is to leverage natural transformations in the unlabeled datasets from multiple devices to generate a supervisory signal for training. In future work, we plan to explore technical solutions to extend \solution{} to audio- and vision- based \problem{} settings. 

\parjump{}
\noindent
We now discuss several limitations of our current approach. 

\parjump{}
\noindent
\textbf{Data Privacy.}  \revisions{\solution{} is designed as a collaborative learning framework, in that it requires raw data from multiple devices to train the HAR model. In practice, the sensor devices owned by a user may be from different device manufacturers, who may not be willing to offload the raw sensor data to a centralized cloud server due to privacy and commercial reasons. We envision two potential solutions to this issue: firstly, model training can be done on a trusted edge device such as a home router and it ensures that a user's data never leaves their premises. Alternatively, federated self-supervised learning approaches~\cite{shi2021federated} can be explored wherein the feature extractor is trained locally on each device and only the gradients of the feature extractor are shared to a central server for aggregation.}

\parjump{}
\noindent
\textbf{Extension to newer SSL algorithms.} \revisions{This work focuses on using a contrastive learning paradigm with positive and negative samples for self-supervised learning. However, recently novel SSL methods have been proposed which do not require negative samples~\cite{chen2021exploring, grill2020bootstrap} and outperform contrastive learning methods such as SimCLR. As a future work, we plan to explore such methods for self-supervised learning in \problem{} settings.}

\section{Conclusion}

We presented Collaborative Self-Supervised Learning (\solution{}), a new method to leverage unlabeled inertial data collected from multiple body-worn devices to learn a good representation of the data. In doing so, we exploited an important characteristic of the \problem{} setting that the time-aligned data from different devices can be considered as natural transformations of each other. Based on this observation, we presented a contrastive learning pipeline which intelligently gathers positive and negative samples from multiple devices, and contrasts them against a sample from the anchor device to generate a supervisory signal from unlabeled data. 

Our key findings are that \solution{} outperforms both fully-supervised and semi-supervised learning techniques in majority of the experiment settings. Secondly, \solution{} is data-efficient and can outperform the fully-supervised baselines using one-tenth of the labeled data in most settings. We also showed that \solution{} can learn well-separable features from the data, and threw light on how it makes its predictions by visualizing saliency maps. Even though our focus in this paper was on the task of human-activity recognition with inertial sensors, our idea of Collaborative Self-Supervised Learning and our proposed framework are general, and they can be applied to other sensing modalities and applications in future work. 

\bibliographystyle{ACM-Reference-Format}
\bibliography{mdfl}


\end{document}